\documentclass[twoside,11pt]{article}
\usepackage{jmlr2e}

\ShortHeadings{Greedy Feature Selection for Subspace Clustering}{Dyer, Sankaranarayanan, Baraniuk}
\firstpageno{1}

\usepackage{xspace}
\usepackage{float}
\usepackage{multirow}
\usepackage{verbatim}
\usepackage{algorithmic, algorithm}
\usepackage{amsbsy,amscd,amsfonts,amsgen,amsmath,amsopn,amssymb,amstext,amsxtra}
\graphicspath{{FIGS/}}

\def \sub {\mathcal{S}}
\def \set {\mathcal{Y}}

\newcommand{{\x}}{\bf{x}}
\newcommand{{\bhat}}{\hat{\bf{b}}}
\newcommand{{\xhat}}{\hat{\bf{x}}}
\newcommand{{\hhat}}{\hat{\bf{h}}}
\newcommand{{\zhat}}{\hat{\bf{z}}}

\newcommand{\atom}{{\bf{\varphi}}}

\newcommand{\dct}{\Phi}

\newcommand{\R}{\mathbb{R}}

\newcommand{{\coef}}{ a}

\newtheorem{THEO}{Theorem}
\newtheorem{LEMM}{Lemma}

\newtheorem{DEFI}{Definition}

\newtheorem{COROLLARY}{Corollary}

\begin{document}

\title{Greedy Feature Selection for Subspace Clustering}

\author{\name Eva L. Dyer \email e.dyer@rice.edu \\
       \addr Department of Electrical \& Computer Engineering\\
       Rice University, Houston, TX, 77005, USA
       \AND
       \name Aswin C. Sankaranarayanan \email saswin@ece.cmu.edu \\
       \addr Department of Electrical \& Computer Engineering\\
       Carnegie Mellon University, Pittsburgh, PA, 15213, USA
       \AND
       \name Richard G. Baraniuk \email richb@rice.edu\\
       \addr Department of Electrical \& Computer Engineering\\
       Rice University, Houston, TX, 77005, USA   
       }

\editor{}

\maketitle

\begin{abstract}
\sloppy{Unions of subspaces provide a powerful generalization to linear subspace models for collections of high-dimensional data. To learn a union of subspaces from a collection of data, sets of signals in the collection that belong to the same subspace must be identified in order to obtain accurate estimates of the subspace structures present in the data. Recently, sparse recovery methods have been shown to provide a provable and robust strategy for {\em exact feature selection} (EFS)---recovering subsets of points from the ensemble that live in the same subspace. In parallel with recent studies of EFS with $\ell_1$-minimization, in this paper, we develop sufficient conditions for EFS with a greedy method for sparse signal recovery known as orthogonal matching pursuit (OMP). Following our analysis, we provide an empirical study of feature selection strategies for signals living on unions of subspaces and characterize the gap between sparse recovery methods and nearest neighbor (NN)-based approaches. In particular, we demonstrate that sparse recovery methods provide significant advantages over NN methods and the gap between the two approaches is particularly pronounced when the sampling of subspaces in the dataset is sparse. Our results suggest that OMP may be employed to reliably recover exact feature sets in a number of regimes where NN approaches fail to reveal the subspace membership of points in the ensemble.}

\end{abstract}




\section{Introduction}
\subsection{Unions of Subspaces}

\sloppy{ With the emergence of novel sensing systems capable of acquiring data at scales ranging from the nano to the peta, modern sensor and imaging data are becoming increasingly high-dimensional and heterogeneous. To cope with this explosion of high-dimensional data, one must exploit the fact that low-dimensional geometric structure exists amongst collections of data.

Linear subspace models are one of the most widely used signal models for collections of high-dimensional data, with applications throughout signal processing, machine learning, and the computational sciences. This is due in part to the simplicity of linear models but also due to the fact that principal components analysis (PCA) provides a closed-form and computationally efficient solution to the problem of finding an optimal low-rank approximation to a collection of data (an ensemble of signals in $\R^n$). More formally, if we stack a collection of $d$ vectors (points) in $\R^n$ into the columns of $Y \in \R^{n \times d}$, then PCA finds the best rank-$k$ estimate of $Y$ by solving
\begin{equation}
\label{eq:pca}
({\rm PCA}) \qquad \min_{X \in {\mathbb R}^{n \times d}} \quad \| Y - X \|_F \quad {\rm subject~ to} \quad {\rm rank}(X) \le k.
\end{equation}

In many cases, a linear subspace model is sufficient to characterize the intrinsic structure of the ensemble; however, in many emerging applications, a single subspace is not enough. Instead, ensembles must be modeled as living on a {\em union of subspaces} or a union of affine planes of mixed or equal dimension. Ensembles ranging from collections of images taken of objects under different illumination conditions \citep{basrijacobs,facesubs}, motion trajectories of point-correspondences \citep{kanatani01}, to structured sparse and block-sparse signals \citep{unionsubs,Blu,modelcs} are all well-approximated by a union of low-dimensional subspaces or a union of affine hyperplanes. Union of subspace models have also found utility in the classification of signals collected from complex and adaptive systems at different instances in time, e.g., electrical signals collected from the brain's motor cortex \citep{gowree2011neuraldecode}. 


Unions of subspaces provide a natural extension to single subspace models, but providing an extension of PCA that leads to {\em provable} guarantees for learning multiple subspaces is challenging. This is due to the fact that segmentation---the identification of points that live in the same subspace---and subspace estimation must be performed simultaneously \citep{GPCA2005,vidal2011spmag}. However, if we can accurately sift through the points in the ensemble and determine which points lie along or near the same subspace, then subspace estimation becomes trivial. For this reason, many state-of-the art methods for learning unions of subspaces rely on first forming {\em local subspace estimates}{\footnote{A local subspace estimate is a low-rank approximation formed from a subset of points in the ensemble, rather than from the entire collection of data.}} from a subset of points in the data \citep{vidal2011spmag,vidaljournal}. 

A common heuristic used to obtain local subspace estimates is to select points that lie within an Euclidean neighborhood of one another (or a fixed number of nearest neighbors (NNs)) and then form a local estimate from the set of NNs. At a high-level, most NN-based approaches for subspace clustering can be summarized as consisting roughly of the following three steps:
\begin{itemize}
\item[(1)] For the $i^{\rm th}$ point in the set, $y_i$, select a set of points from the ensemble that live within an $\epsilon$-radius from $y_i$ in terms of their Euclidean distance. Denote this subset of points $Y_{\Lambda}$, where $\Lambda$ is an index set containing the indices of all the neighbors of $y_i$. 
\item[(2)] Form a low-rank PCA estimate by solving (\ref{eq:pca}) for the points in the sub-matrix $Y_{\Lambda}$.
\item[(3)] Compute the {\em subspace affinity matrix} $W \in \R^{d \times d}$ for the ensemble, where the $(i,j)$ entry of the matrix represents the likelihood that $y_i$ and $y_j$ live close to the same subspace or whether $y_i$ and $y_j$ produce similar local subspace estimates. 
\end{itemize}
Methods that use NN sets to form local subspace estimates from the data include local subspace affinity (LSA) \citep{LSA}, spectral clustering based on locally linear approximations \citep{locallinear}, spectral curvature clustering \citep{chen2009scc}, and local best-fit flats \citep{zhang2010hybrid}. The main differences between these methods lie either in the way that the entries of the affinity matrix are computed in Step 3 or the way in which this matrix is used to obtain an estimate of the underlying subspace structures present in the ensemble. In the case of approaches built upon spectral clustering \citep{graphcuts,ng2002spectral}, one performs spectral clustering on the subspace affinity matrix for the ensemble in order to cluster the data into different subspaces. In the case of consensus-based approaches, one finds a robust estimate of the mode of the local subspace estimates formed in Step 2; this problem can also be posed as an optimization on the subspace affinity matrix. We point the reader to \citep{vidal2011spmag} for a thorough review of methods for subspace clustering.

When the subspaces present in the ensemble are linearly separable or non-intersecting, local subspace estimates formed from NNs provide relatively reliable and stable estimates of the subspaces present in the ensemble. However, neighborhood-based approaches quickly fail as the separation between the two structures decreases and as the subspace dimension increases relative to the number of points in each subspace. This is due in part to the fact that, as the dimension of the intersection between two subspaces increases, the Euclidean distance between points becomes a poor predictor of which points belong to the same subspace. Thus, we seek an alternative strategy for forming a local estimate that does not rely solely on whether points in the same subspace live in a local Euclidean neighborhood. Instead, our goal is to identify another strategy for ``feature selection'' that returns sets of points (feature sets) that lie along the same subspace.

\subsection{Exact Feature Selection}
Instead of computing local subspace estimates from sets of NNs,  \cite{elhamifar2009sparse} propose a novel approach for feature selection based upon forming sparse representations of the data via $\ell_1$-minimization. The main intuition underlying their approach is that when a sparse representation of a point is formed with respect to the remaining points in the ensemble, the representation should only consist of other points that belong to the same subspace. When a sparse representation consists of points that lie in the same subspace, we say that  {\em exact feature selection} (EFS) occurs. Under certain assumptions on both the sampling and ``distance between subspaces'',{\footnote{The distance between a pair of subspaces is typically measured with respect to the principal angles between the subspaces or other related distances on the Grassmanian manifold.}} this approach to feature selection leads to provable guarantees that EFS will occur \citep{vidaldisjoint,Candes12}, even when the subspaces intersect.



We refer to this application of sparse recovery as {\em endogenous sparse recovery} due to the fact that representations are not formed from an external collection of primitives (such as a basis or dictionary) but are formed ``from within'' the data.  Formally, for a set of $d$ signals $ \set = \{y_1, \dots, y_d \}$, each of dimension $n$, the sparsest representation of the $i^{\rm th}$ point $y_i$ is defined as
\begin{equation}
\label{eq:endo}
c_i^{\ast} ~=~ \arg \min_{c \in \R^d}~~ \| c \|_0 \qquad \mathrm{subject~ to} \qquad  y_i = \sum_{j \ne i} c(j) y_j , 
\end{equation}
where $\| c\|_0$ counts the number of non-zeroes in its argument. Let $\Lambda^{(i)} = {\rm supp}(c^{\ast}_i)$ denote the subset of points selected to represent the $i^{\rm th}$ point and $c^{\ast}_i(j)$ denote the contribution of the $j^{\rm th}$ point to the endogenous representation of $y_i$. By penalizing representations that require a large number of non-zero coefficients, the resulting representation will be sparse. 

In general, finding the sparsest representation of a signal has combinatorial complexity; thus, sparse recovery methods such as basis pursuit (BP) \citep{DonohoBP} or low-complexity greedy methods \citep{DavisOMP} are employed to obtain approximate solutions to (\ref{eq:endo}). 



\subsection{Contributions}
In \cite{vidaldisjoint}, the authors show that when subspaces are {\em disjoint} (intersect only at the origin) and the minimum principal angle between subspaces is sufficiently large, the points selected by BP will belong to the same subspace, i.e., EFS is guaranteed. Recently, \cite{Candes12} developed guarantees for EFS with BP from unions of intersecting subspaces.

In parallel with recent developments for subspace clustering with BP, in this paper, we study EFS with a low-complexity and greedy method for sparse signal recovery known as orthogonal matching pursuit (OMP). The main result of our analysis is a new geometric condition (Thm.\ \ref{theo:EFS}) for EFS that highlights the tradeoff between the: {\em mutual coherence} or similarity between points living in different subspaces and the {\em covering radius} of the points within a common subspace. The covering radius can be interpreted as the radius of the largest ball that can be embedded within each subspace without touching a point in the ensemble; the vector that lies at the center of this open ball, or the vector in the subspace that attains the covering radius is referred to as a {\em deep hole}. Thm.\ \ref{theo:EFS} suggests that subspaces can be arbitrarily close to one another and even intersect, as long as the data is distributed ``nicely'' along each subspace. By ``nicely'', we mean that the points that lie on each subspace do not cluster together, leaving large gaps in the sampling of the underlying subspace. In Fig.\ \ref{fig:deephole}, we illustrate the covering radius of a set of points on the sphere (the deep hole is denoted by a star).

\begin{figure}[t!]
\begin{center}
\centerline{\includegraphics[width=3.7in]{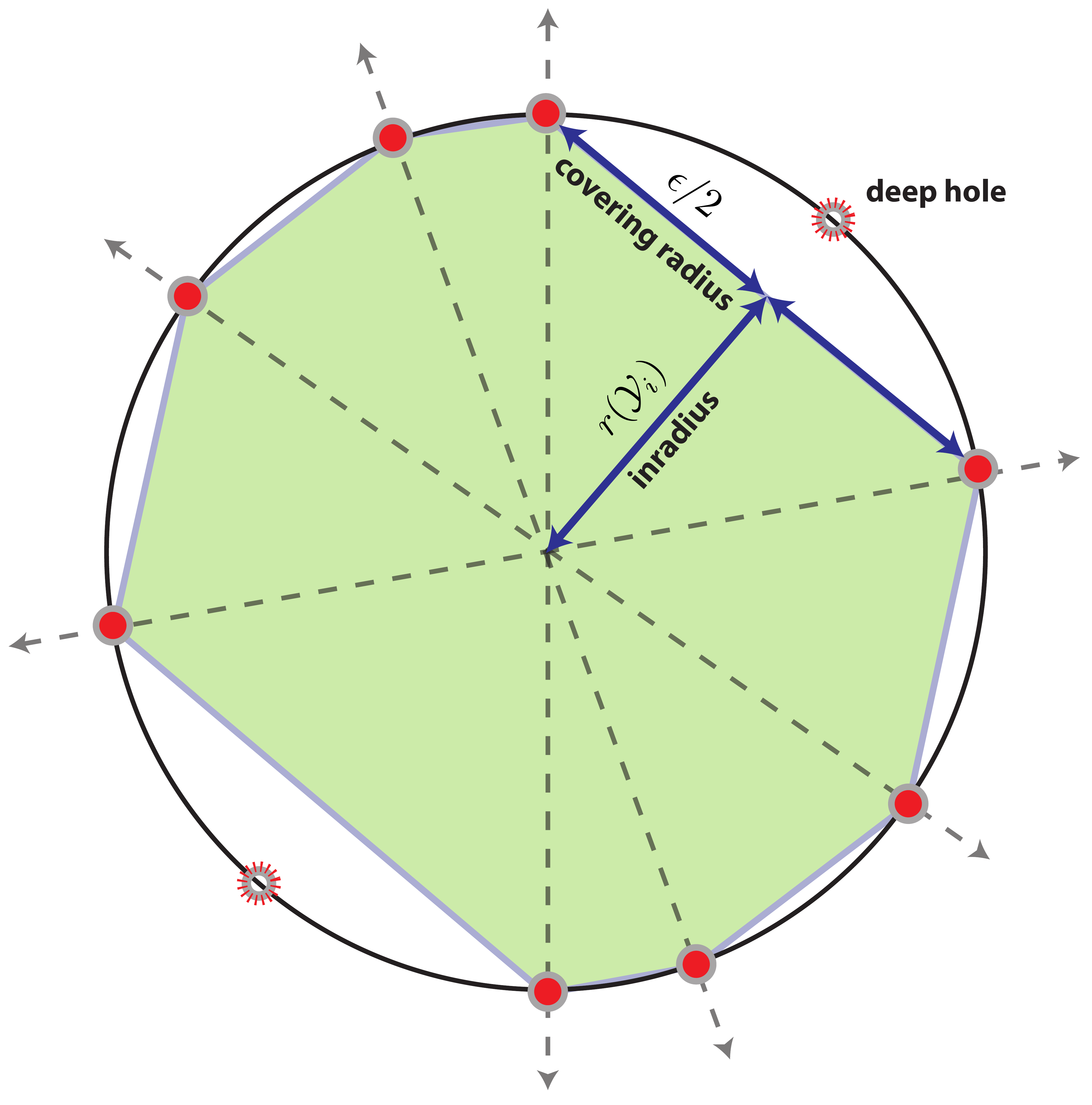}}
\caption[{\em Covering radius and deep holes.}]{ \label{fig:deephole} {\em Covering radius of points in a normalized subspace.} The interior of the antipodal convex hull of points in a normalized subspace---a subspace of $\R^n$ mapped to the unit $\ell_2$-sphere---is shaded. The vector in the normalized subspace (unit circle) that attains the covering radius (deep hole) is marked with a star: when compared with the convex hull, the deep hole coincides with the maximal gap between the convex hull and the set of all vectors that live in the normalized subspace.}
\end{center}
\vspace{-10mm}
\end{figure}

After introducing a general geometric condition for EFS, we extend this analysis to the case where the data live on what we refer to as an {\em uniformly bounded union} of subspaces (Thm. \ref{theo:boundedvecs}). In particular, we show that when the points living in a particular subspace are incoherent with the principal vectors that support pairs of subspaces in the ensemble, EFS can be guaranteed, even when non-trivial intersections exist between subspaces in the ensemble. Our condition for bounded subspaces suggests that, in addition to properties related to the sampling of the subspace, one can characterize the separability of pairs of subspaces by examining the correlation between the dataset and the unique set of principal vectors that support pairs of subspaces in the ensemble. 


In addition to providing a theoretical analysis of EFS with OMP, the other main contribution of this work is revealing the gap between nearest neighbor-based (NN) approaches and sparse recovery methods, i.e., OMP and BP, for feature selection. In both synthetic  and real world experiments, we observe that while both NN and sparse recovery methods have comparable rates of EFS when the subspaces are densely sampled, when the subspaces are sparsely sampled, sparse recovery methods provide significant advantages over NN. These empirical results point to an advantage of forming sparse representations from within the data; sparse recovery methods provide a natural way to reveal the subspace affinity amongst points that might be far away from one another in a Euclidean sense. By exploiting non-local relationships between points, sparse recovery methods are capable of providing reliable subspace estimates with far fewer points than neighborhood-based estimates. See Fig.\ \ref{fig:faces} for an example of the affinity matrices formed from pairs of face subspaces where the goal is to separate points that live on different ``illumination subspaces''.


\begin{figure}[t!]
\vskip 0.2in
\begin{center}
\centerline{\includegraphics[width=5in]{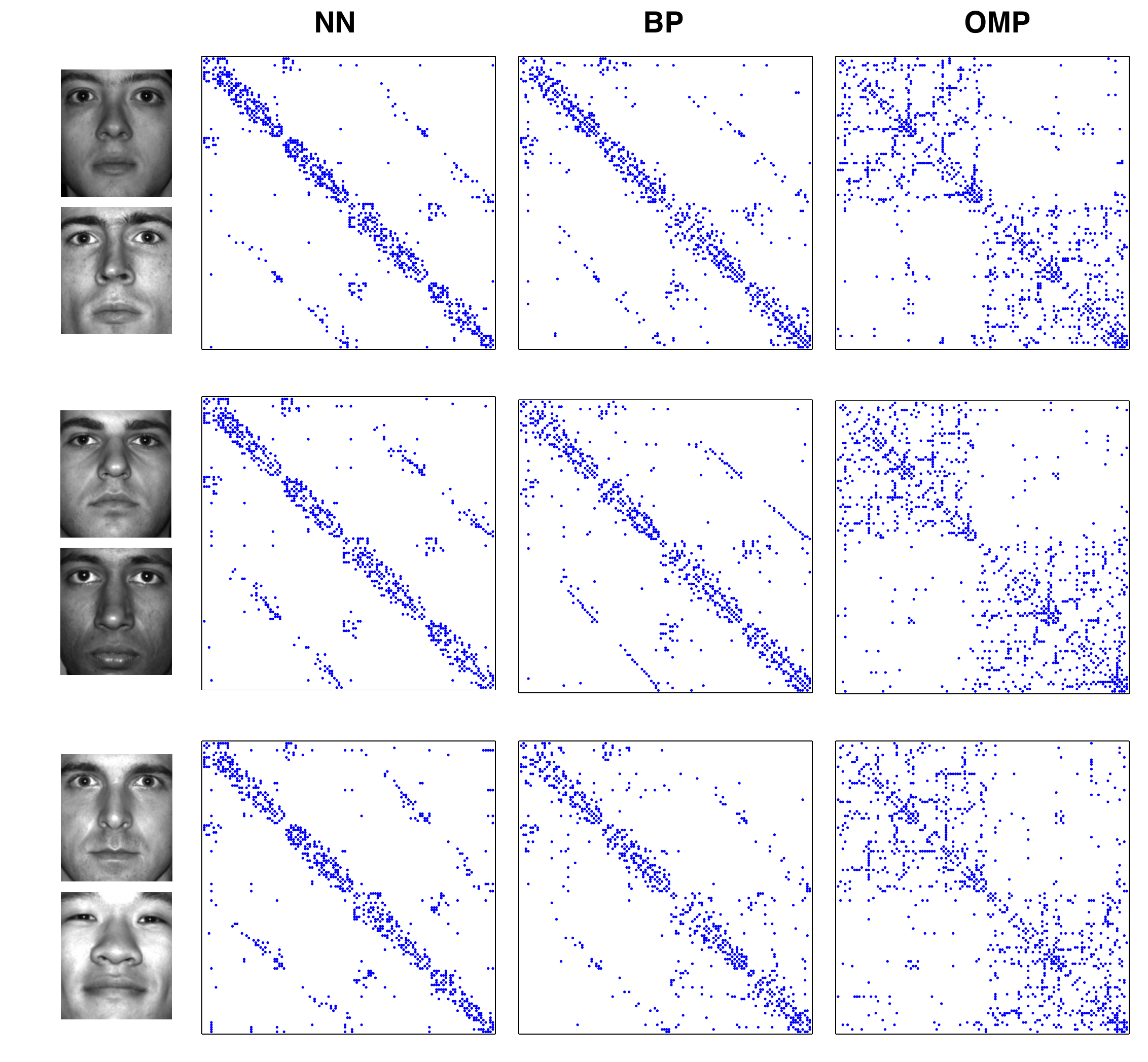}}
\caption[{\em Comparison of subspace affinities.}] {\label{fig:faces} {\em Comparison of subspace affinity matrices for illumination subspaces.}~In each row, we display the subspace affinity matrices obtained for a different pair of illumination subspaces in the dataset, for NN (left), BP (middle), and OMP (right). To the left of the affinity matrices, we display an exemplar image from each illumination subspace.}
\end{center}
\vskip -0.3in
\end{figure}



\subsection{Paper Organization}
We now provide a roadmap for the rest of the paper.
\vspace{3mm}


\noindent {\bf Section 2.} We introduce our signal model, the sparse subspace clustering (SSC) algorithm introduced in \citep{elhamifar2009sparse}, and describe how OMP may be used for feature selection in subspace clustering.\\

\noindent {\bf Section 3 and 4.} We develop the main theoretical results of this paper and provide new geometric insights into EFS from unions of subspaces. We introduce sufficient conditions for EFS to occur with OMP for general unions of subspaces in Thm.\ \ref{theo:EFS}, disjoint unions in Cor.\ \ref{coro:EFSdisjoint}, and uniformly bounded unions in Thm.\ \ref{theo:boundedvecs}.\\


\noindent {\bf Section 5.} We conduct a number of numerical experiments to validate our theory and compare sparse recovery methods with NN-based feature selection. Experiments are provided for both synthetic and real data.\\

\noindent {\bf Section 6.} We discuss the implications of our theoretical analysis and empirical results on sparse approximation, dictionary learning, and compressive sensing. We conclude with a number of interesting open questions and future lines of research.\\

\noindent {\bf Section 7.} We supply the proofs of the results contained in Sections 3 and 4.


\subsection{Notation}
In this paper, we will work solely in real finite-dimensional vector spaces, $\R^n$. We write vectors $x$ in lowercase script, matrices $A$ in uppercase script, and scalar entries of vectors as $x(j)$. The standard $p$-norm is defined as
$$ \| x \|_p = \bigg( \sum_{j =1}^n | x(j) | ^p \bigg)^{1/p},$$ where $p \ge 1$. The ``$\ell_0$-norm'' of a vector $x$ is defined as the number of non-zero elements in $x$. The support of a vector $x$, often written as ${\rm supp}(x)$, is the set containing the indices of its non-zero coefficients; hence, $\| x \|_0 = | {\rm supp}(x)|$. We denote the Moore-Penrose pseudoinverse of a matrix $A$ as $A^{\dagger}$. If $A = U \Sigma V^T$ then $A^{\dagger} = V \Sigma^{+} U^T$, where we obtain $\Sigma^{+}$ by taking the reciprocal of the entries in $\Sigma$, leaving the zeros in their places, and taking the transpose. An orthonormal basis (ONB) $\Phi$ that spans the subspace $\sub$ of dimension $k$ satisfies the following two properties: $\Phi^T \Phi = I_k$ and ${\rm range}(\Phi) = \sub$, where $I_k$ is the ${k \times k}$ identity matrix. Let  $P_{\Lambda} = X_{\Lambda} X^{\dagger}_{\Lambda}$ denote an ortho-projector onto the subspace spanned by the sub-matrix $X_{\Lambda}$.

\section{Greedy Feature Selection for Subspace Clustering}
In this section, we introduce our signal model, detail the sparse subspace clustering (SSC) method developed by \cite{elhamifar2009sparse}, and discuss the use of orthogonal matching pursuit for feature selection in subspace clustering.

\subsection{Signal Model}

Given a set of $p$ subspaces of $\R^n$, $\{\sub_1, \dots, \sub_p\}$, each of dimension $k_i \le k$, we generate a ``subspace cluster'' by sampling $d_i$ points from the $i^{\rm th}$ subspace $\sub_i$. Let $\widetilde{\set}_i $ denote the set of points in the $i^{\rm th}$ subspace cluster and let $\widetilde{\set}= \cup_{i=1}^p \widetilde{\set}_i$ denote the union of these $p$ subspace clusters. Each point in $\widetilde{\set}$ is mapped to the unit sphere to generate a union of normalized subspace clusters. Let $$\set = \bigg\{ \frac{y_1} {\|y_1\|_2}, \frac{y_2}{\|y_2\|_2}, \cdots, \frac{y_d}{\|y_d\|_2}  \bigg \}$$ denote the resulting set of unit norm points and let $\set_i$ be the set of unit norm points that lie in the span of subspace $\sub_i$. Let $\set_{ - i} = \set \setminus \set_i$ denote the set of points in $\set$ with the points in $\set_i$ excluded.

Let $Y = [ Y_1 ~ Y_2 ~  \cdots~ Y_p]$ denote the matrix of normalized data, where each point in $\set_i$ is stacked into the columns of $Y_i \in \R^{n \times d_i}$. The points in $Y_i$ can be expanded in terms of an ONB $\dct_i \in \R^{n \times k_i}$ that spans $\sub_i$ and subspace coefficients $A_i = \dct_i^T Y_i$, where $Y_i = \dct_i A_i$. Let $Y_{ - i}$ denote the matrix containing the points in $Y$ with the submatrix $Y_i$ excluded.

\subsection{Sparse Subspace Clustering}
In \citep{elhamifar2009sparse}, the authors propose a novel approach to subspace clustering called sparse subspace clustering (SSC) that employs a relaxation of the $\ell_0$-minimization problem in (\ref{eq:endo}). To be precise, the SSC algorithm proceeds by solving the following basis pursuit (BP) problem \citep{DonohoBP} for each point in $\mathcal{Y}$:
\begin{equation}
\label{eq:ssc}
c_i^{\ast} ~=~ \arg ~\min_{c \in {\mathbb R}^d} ~~  \| c \|_1 \qquad \mathrm{subject~ to} \qquad  y_i = \sum_{j \ne i} c(j) y_j.
\end{equation}
After finding the solution to this $\ell_1$-minimization problem for each point in the ensemble, each $d$-dimensional feature vector $c_i^{\ast}$ is placed into the $i^{\rm th}$ row or column of $C \in \R^{d \times d}$ and spectral clustering \citep{graphcuts,ng2002spectral} is performed on the graph Laplacian of the affinity matrix $W = |C| + |C^T |$. 

In \citep{vidaljournal}, the authors provide an extension of SSC to the case where the data might not admit an exact representation with respect to other points in the ensemble. In this case, they employ an inequality constrained version of BP known as basis pursuit denoising (BPDN) for feature selection; for each point $y_i$, they solve the following problem
\begin{equation}
\label{eq:bpdn}
c_i^{\ast} ~=~ \arg \min_{c \in {\mathbb R}^d} ~~ \| c \|_1\qquad \mathrm{subject~ to} \qquad \|  y_i - \sum_{j \ne i} c(j) y_j\|_2 < \kappa,
\end{equation}
where $\kappa$ is a parameter that is selected based upon the amount of noise in the data. Recently, \cite{Xu2013} provided an analysis of EFS for a variant of the formulation in (\ref{eq:bpdn}) for noisy unions of subspaces. In \citep{Candes13}, the authors propose a robust procedure for subspace clustering from noisy data that extends the BPDN framework studied in \citep{vidaljournal,Xu2013}.

\subsection{Greedy Feature Selection}
\label{sec:ompalg}
Instead of solving the sparse recovery problem in (\ref{eq:endo}) via $\ell_1$-minimization, as originally proposed in SSC, we will study the behavior of a low-complexity method for sparse feature selection known as orthogonal matching pursuit (OMP). We detail the OMP algorithm in Alg.\ \ref{alg:omp}. 

For each point $y_i$, we solve Alg. \ref{alg:omp} to obtain a $k$-sparse representation of the signal with respect to the remaining points in $Y$. The output of the OMP algorithm is a {\em feature set}, $\Lambda^{(i)}$, which indexes the columns in $Y$ selected to form an endogenous representation of $y_i$. After computing feature sets for each point in the dataset, either a spectral clustering method or a consensus-based method may then be employed. In \citep{DyerMS}, we introduce a consensus-based algorithm for subspace clustering that uses OMP for feature selection and also provide an empirical study of different feature selection strategies for both consensus and spectral clustering-based approaches on both synthetic and real world data (Chap. 7--8).

To form the subspace affinity for the dataset, a $d$-dimensional sparse feature vector $\bar{c}_i$ is computed by stacking the $k$-dimensional projection $c_i = Y_{\Lambda^{(i)}}^{\dagger} y_i$ into the entries of $\bar{c}_i$ indexed by the feature set $\Lambda^{(i)}$, where $Y_{\Lambda^{(i)}}^{\dagger} \in \R^{k \times n}$ is the pseudoinverse of the submatrix $Y_{\Lambda^{(i)}} \in \R^{n \times k}$. The remaining entries in $\bar{c}_i$ are set to zero. The feature vector $\bar{c_i}$ is then stacked in the $i^{\rm th}$ row of $C \in \R^{d \times d}$ and the subspace affinity matrix of the ensemble is computed as $W = |C| + |C^T|$.  Finally, spectral clustering is performed either on the graph Laplacian or the normalized graph Laplacian of $W$.

Although OMP is known to be suboptimal for standard applications of sparse signal recovery, our empirical results provided in Section \ref{sec:facesubs} suggest that OMP provides a low-complexity alternative to $\ell_1$-minimization methods for SSC. An obvious advantage of using greedy methods is that they exhibit reduced computational complexity when compared to convex optimization-based approaches, thus enabling their use for feature selection from large collections of data. In addition, we find that despite the fact that BPDN provides better rates of EFS than OMP when we carefully tune the noise parameter $\kappa$, OMP provides comparable (and in some cases better) clustering performance than BPDN for the same choice of $\kappa$. These empirical results suggest that OMP offers a powerful low-complexity alternative to $\ell_1$-minimization for feature selection in SSC. We point the reader to Fig.\ \ref{fig:faces} for an example of the affinity matrices obtained via OMP, BP, and NN for collections of images of faces under different lighting conditions.

\begin{algorithm}[t!]
   \caption{: Orthogonal Matching Pursuit}
   \label{alg:omp}
\begin{algorithmic}
   \STATE {\bfseries Input:} Input signal $y \in \R^n$, a matrix $A \in \R^{n \times d}$ containing $d$ signals $\{ a_i \}_{i=1}^d$ in its columns, and a stopping criterion (either the sparsity $k$ or the approximation error $\kappa$).\\
   \STATE {\bfseries Output:} An index set $\Lambda$ containing the indices of all atoms selected in the pursuit.
\STATE{\bfseries Initialize:} Set the residual to the input signal $s = y$.
\STATE{1. Select the atom that is maximally correlated with the residual and add it to $\Lambda$ 
$$\Lambda \leftarrow \Lambda ~ \cup ~ \arg ~ \max_{i}  | \langle a_i, s \rangle | .$$}
\STATE{2. Update the residual by projecting $s$ into the space orthogonal to the span of $A_{\Lambda}$ $$s \leftarrow (I - A_{\Lambda} A^{\dagger}_{\Lambda}) y.$$}
\STATE{3. Repeat steps (1)--(2) until the stopping criterion is reached, e.g., either $| \Lambda | = k$ or the norm of the residual $\|s\| \le \kappa$.}
\end{algorithmic}
\end{algorithm}

\section{Exact Feature Selection from Unions of Subspaces}
\label{sec:efs_union}
In this section, we provide a formal definition of EFS and develop sufficient conditions that guarantee that EFS will occur for all of the points contained within a particular subspace cluster. 

\subsection{Exact  Feature Selection}
In order to guarantee that OMP returns a sample set that yields an accurate local subspace estimate, we will be interested in determining when the feature set returned by Alg.\ \ref{alg:omp} only contains points that belong to the same subspace cluster; in this case, we say that {\em exact feature selection} (EFS) occurred. EFS provides a natural condition for studying performance of both subspace consensus and spectral clustering methods due to the fact that when EFS occurs for a point, this results in a local subspace estimate that coincides with one of the true subspaces contained within the data. We now supply a formal definition of EFS.
\begin{DEFI}[Exact feature selection] Let $\set_k = \{ y : (I - P_k) y = 0 , ~ y \in \mathcal{Y}  \}$ index the set of points in $\mathcal{Y}$ that live in the span of subspace $\sub_k$, where $P_k$ is a projector onto the span of subspace $\sub_k$. For a point $y \in \set_k$ with feature set  $\Lambda$, if $y_i \subseteq \set_k, ~\forall i \in \Lambda$, we say that $\Lambda$ contains exact features. 
\end{DEFI}


\subsection{Geometric Conditions for EFS}
\label{sec:EFSanalysis}


\subsubsection{Preliminaries}
Our main result in Thm.\ \ref{theo:EFS} below requires measures of both the distance between points in {\em different subspace clusters} and within the {\em same subspace cluster}. A natural measure of the similarity between points living in different subspaces is the {\em mutual coherence}. A formal definition of the mutual coherence is provided below in Def. \ref{def:mutualcoh}. 
\begin{DEFI}
\label{def:mutualcoh}
The mutual coherence between the points in the sets $(\set_i, \set_j)$ is defined as
\begin{equation}
\mu_c(\set_i,\set_j) ~ = ~ \max_{u \in \set_i, v \in \set_j} ~~ | \langle u , v \rangle |, ~~ {\rm where}~~ \| u \|_2 = \| v \|_2 = 1.
\end{equation}
\end{DEFI}
In words, the mutual coherence provides a point-wise measure of the normalized inner product (coherence) between all pairs of points that lie in two different subspace clusters. 

Let $\mu_c(\set_i)$ denote the maximum mutual coherence between the points in $\set_i$ and all other subspace clusters in the ensemble, where $$ \mu_c(\set_i) = \max_{j \ne i}~ \mu_c(\set_i,\set_j).$$

A related quantity that provides an upper bound on the mutual coherence is the cosine of the first {\em principal angle} between the subspaces. The first principal angle $\theta_{ij}^{\ast}$ between subspaces $\sub_i$ and $\sub_j$, is the smallest angle between a pair of unit vectors $(u_1,v_1)$ drawn from $\sub_i \times \sub_j$. Formally, the first principal angle is defined as
\begin{equation}
\label{defi:minangle}
\theta_{ij}^{\ast} = \min_{u \in \sub_i, ~v \in \sub_j} ~~ \arccos~ \langle u, v \rangle \quad {\rm subject~ to} \quad  \| u \|_2 = 1, \| v \|_2 = 1.
\end{equation}

Whereas the mutual coherence provides a measure of the similarity between a pair of unit norm vectors that are contained in the sets $\set_i$ and $\set_j$, the cosine of the minimum principal angle provides a measure of the similarity between a pair of unit norm vectors that lie in the span of $ \sub_i \times \sub_j$. For this reason, the cosine of the first principal angle provides an upper bound on the mutual coherence. The following upper bound is in effect for each pair of subspace clusters in the ensemble:
\begin{equation} 
\label{eq:cosangle}
\mu_c(\set_i,\set_j) \le \cos (\theta_{ij}^{\ast}).
\end{equation}

To measure how well points in the same subspace cluster cover the subspace they live on, we will study the covering radius of each normalized subspace cluster relative to the projective distance. Formally, the covering radius of the set $\set_k$ is defined as
\begin{equation}
\label{eq:cover}
{\rm cover}(\set_k)~=~\max_{u \in \sub_k}~~ \min_{y \in \set_k} ~~  {\rm dist}(u , y),
\end{equation}
where the projective distance between two vectors $u$ and $y$ is defined relative to the acute angle between the vectors
\begin{equation}
\label{eq:projdist}
{\rm dist}(u,y) = \sqrt{1 - {\frac{| \langle u, y \rangle |^2}{\|u\|_2 \|y \|_2}} }.
\end{equation}
The covering radius of the normalized subspace cluster $\set_i$ can be interpreted as the size of the largest open ball that can be placed in the set of all unit norm vectors that lie in the span of $\sub_i$, without touching a point in $\set_i$. 

Let $(u_i^{\ast}, y_i^{\ast})$ denote a pair of points that attain the maximum covering diameter for $\set_i$; $u_i^{\ast} \in \sub_i$ is referred to as a deep hole in $\set_i$ along $\sub_i$. The covering radius can be interpreted as the sine of the angle between the deep hole $u_i^{\ast}$ and its nearest neighbor $y_i^{\ast} \in \set_i$. We show the geometry underlying the covering radius in Fig. \ref{fig:deephole}.


In the sequel, we will be interested in the maximum (worst-case) covering attained over all $d_i$ sets formed by removing a single point from $\set_i$. We supply a formal definition below in Def. \ref{def:cover}. 
\begin{DEFI}
\label{def:cover}
The maximum covering diameter $\epsilon$ of the set $\set_i$ along the subspace $\sub_i$ is defined as
\begin{equation*}
\epsilon ~=~ \max_{j = 1, \dots, d_i}~~ 2~ {\rm cover}( \{ \set_i \setminus y_j \}).
\end{equation*}
Hence, the covering radius equals $\epsilon/2$.
\end{DEFI}

A related quantity is the {\em inradius} of the set $\set_i$, or the cosine of the angle between a point in $\set_i$ and any point in $\sub_i$ that attains the covering radius. The relationship between the covering diameter $\epsilon$ and inradius $r(\set_i)$ is given by
\begin{equation}
 \label{eq:inradius} 
 r(\set_i) = \sqrt{1 - \frac{\epsilon^2}{4}}.
 \end{equation}
A geometric interpretation of the inradius is that it measures the distance from the origin to the maximal gap in the antipodal convex hull of the points in $\set_i$. The geometry underlying the covering radius and the inradius is displayed in Fig. \ref{fig:deephole}. 

\subsubsection{General Result for EFS}
We are now equipped to state our main geometric result for EFS with OMP. The proof is contained in Section \ref{sec:proofefstheo}.
\begin{THEO} \label{theo:EFS} Let $\epsilon$ denote the maximal covering diameter of the subspace cluster $\set_i$ as defined in Def.\ \ref{def:cover}. A sufficient condition for EFS to occur for all points in $\set_i$ is that the mutual coherence
\begin{equation} 
\label{eq:efs}
\mu_c(\set_i)  ~ < ~ \sqrt{1 - \frac{\epsilon^2}{4}} - \frac{\epsilon}{\sqrt[4]{12}} \max_{j \ne i} \cos(\theta_{ij}^{\ast}) ,
\end{equation} 
where $\theta_{ij}^{\ast}$ is the minimum principal angle defined in (\ref{defi:minangle}). 
\end{THEO}
In words, this condition requires that the mutual coherence between points in {\em different subspaces} is less than the difference of two terms that both depend on the covering radius of points along a {\em single subspace}. The first term on the RHS of (\ref{eq:efs}) is equal to the inradius, as defined in (\ref{eq:inradius}); the inradius provides a measure of the coherence a points that attains the covering radius of the subspace cluster and its nearest neighbor in $\set_i$. The second term on the RHS of (\ref{eq:efs}) is the product of the cosine of the minimum principal angle between pairs of subspaces in the ensemble and the covering diameter $\epsilon$ of the points in $\set_i$.

When subspaces in the ensemble intersect, i.e., $\cos(\theta_{ij}^{\ast}) = 1$, condition (\ref{eq:efs}) in Thm.\ \ref{theo:EFS} can be simplified to
\begin{equation} \label{eq:efssimple} \mu_c(\set_i)  < \sqrt{1 - \frac{\epsilon^2}{4}} - \frac{\epsilon}{{\sqrt[4]{12}}} \approx \sqrt{1 - \frac{\epsilon^2}{4}} - \frac{\epsilon}{1.86}.
\end{equation}
In this case, EFS can be guaranteed as long as the points in different subspace clusters are bounded away from intersections between subspaces. When the covering radius shrinks to zero, Thm.\ \ref{theo:EFS} requires that $\mu_c < 1$, or that points from different subspaces do not lie exactly in the subspace intersection, i.e., are identifiable from one another. 



\subsubsection{Geometry underlying EFS}
\label{sec:geoanalysis}
The main idea underlying the proof of Thm.\ \ref{theo:EFS} is that, at each iteration of Alg.\ \ref{alg:omp}, we require that the residual used to select a point to be included in the feature set is closer to a point in the {\em correct subspace cluster} ($\set_i$) than a point in an {\em incorrect subspace cluster} ($\set_{-i}$). To be precise, we require that the normalized inner product of the residual signal $s$ and all points outside of the correct subspace cluster
\begin{equation}
\label{eq:select}
\max_{y \in \set_{-i}} \frac{ |\langle s, y \rangle |}{\| s \|_2}  < r(\set_i),
\end{equation}
at each iteration of Alg.\ \ref{alg:omp}. To provide the result in Thm.\ \ref{theo:EFS}, we require that (\ref{eq:select}) holds for all $s \in \sub_i$, or all possible residual vectors.

A geometric interpretation of the EFS condition in Thm.\ \ref{theo:EFS} is that the orthogonal projection of all points outside of a subspace must lie within the antipodal convex hull of the set of normalized points that span the subspace. To see this, consider the projection of the points in $\set_{-i}$ onto $\sub_i$. Let $z_j^{\ast}$ denote the point on subspace $\sub_i$ that is closest to the signal $y_j \in \set_{-i}$,
$$ z_j^{\ast} =  \arg \min_{z \in \sub_i}  ~  \| z - y_j \|_2.$$ 
We can also write this projection in terms of a orthogonal projection operator $P_i = \dct_i \dct_i^T$, where $\dct_i$ is an ONB that spans $\sub_i$ and $z_j^{\ast} = P_i y_j$.

\begin{figure}[t!]
\begin{center}
\centerline{\includegraphics[width=6in]{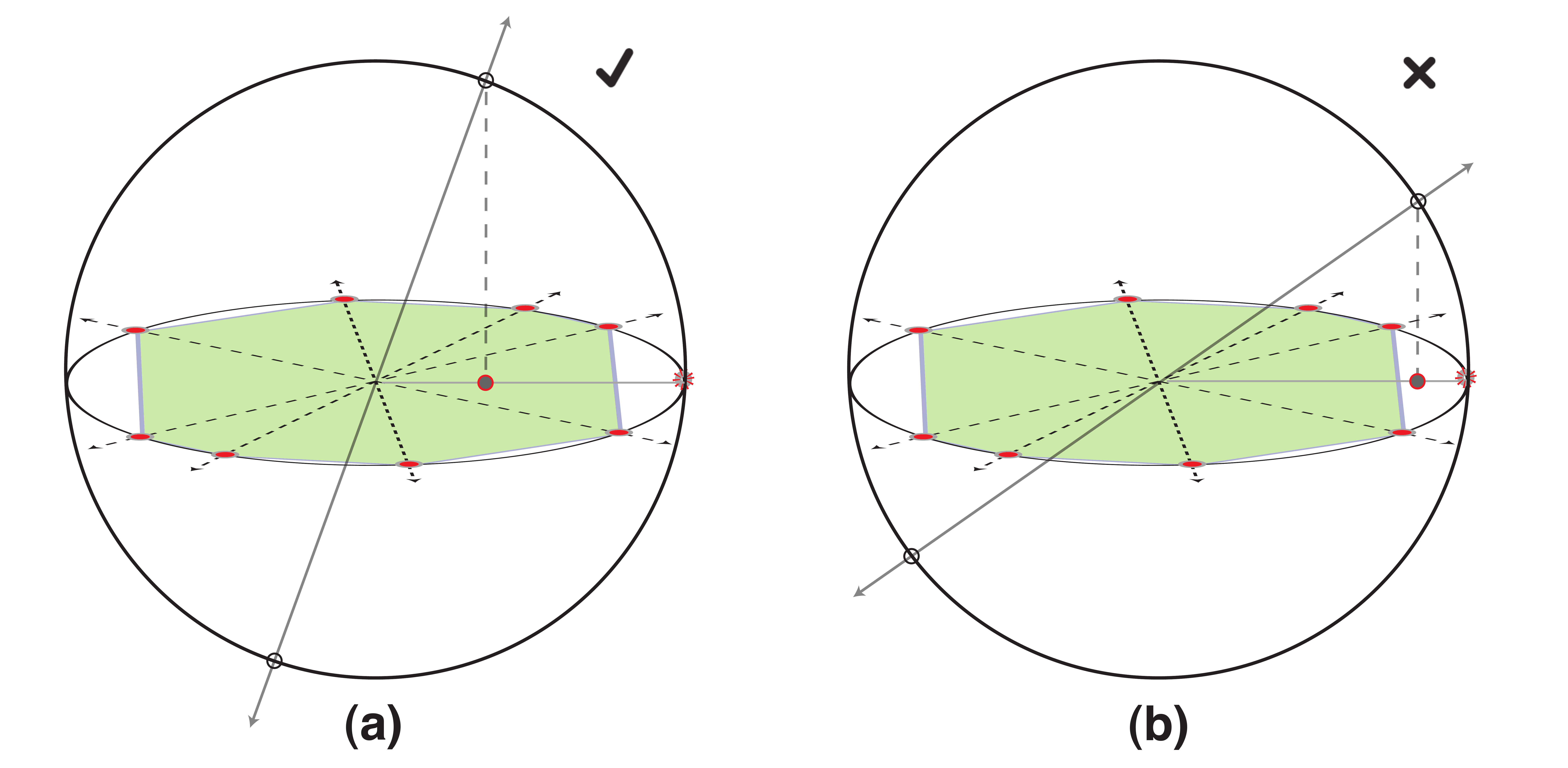}}
\caption[ {\em Geometry underlying EFS from disjoint subspaces.}]{ \label{fig:3Dgeo} {\em Geometry underlying EFS.} A union of two disjoint subspaces of different dimension: the antipodal convex hull of a set of points (red circles) living on a 2D subspace is shaded (green). In (a), we show an example where EFS is guaranteed---the projection of points along the 1D subspace lie inside the shaded convex hull of points in the plane. In (b), we show an example where EFS is not guaranteed---the projection of points along the 1D subspace lie outside the shaded convex hull.}
\end{center}
\vspace{-4mm}
\end{figure}

By definition, the normalized inner product of the residual with points in incorrect subspace clusters is upper bounded as
\begin{align}
\max_{y_j \in \set_{-i}} \frac{|\langle s, y_j \rangle |}{\| s \|_2}  & \le  \max_{y_j \in \set_{-i}} \frac{|\langle z_j^{\ast}, y_j \rangle |}{\| z_j^{\ast} \|_2} = \max_{y_j \in \set_{-i}} \cos\angle \{ z_j^{\ast},y_j \}
\end{align}
Thus to guarantee EFS, we require that the cosine of the angle between all signals in $\set_{-i}$ and their projection onto $\sub_i$ is less than the inradius of $\set_i$. Said another way, the EFS condition requires that the length of all projected points be less than the inradius of $\set_i$. 

In Fig.\ \ref{fig:3Dgeo}, we provide a geometric visualization of the EFS condition for a union of disjoint subspaces (union of a 1D subspace with a 2D subspace). In (a), we show an example where EFS is guaranteed because the projection of the points outside of the 2D subspace lie well within the antipodal convex hull of the points along the normalized 2D subspace (ring). In (b), we show an example where EFS is not guaranteed because the projection of the points outside of the 2D subspace lie outside of the antipodal convex hull of the points along the normalized 2D subspace (ring).

\subsubsection{EFS for Disjoint Subspaces}
When the subspaces in the ensemble are {\em disjoint}, i.e., $\cos(\theta_{ij}^{\ast}) < 1$, Thm.\ \ref{theo:EFS} can be simplified further by using the bound for the mutual coherence in (\ref{eq:cosangle}). This simplification results in the following corollary.
 \begin{COROLLARY} \label{coro:EFSdisjoint}Let $\theta_{ij}^{\ast}$ denote the first principal angle between disjoint subspaces $\sub_i$ and $\sub_j$, and let $\epsilon$ denote the maximal covering diameter of the points in $\set_i$. A sufficient condition for EFS to occur for all points in $\set_i$ is that 
\begin{equation}
\max_{j \ne i} ~ \cos(\theta_{ij}^{\ast}) ~ < ~ \frac{ \sqrt{1 - \epsilon^2/4} }{ 1 + \epsilon/\sqrt[4]{12} }.
\end{equation}
\end{COROLLARY}

\subsection{Connections to Previous Work}
\label{sec:prevwork}
In this section, we will connect our results for OMP with previous analyses of EFS with BP provided in \citep{vidaldisjoint,Candes12} for disjoint and intersecting subspaces respectively. Following this, we will contrast the geometry underlying EFS with exact recovery conditions used to guarantee support recovery for both OMP and BP \citep{Tropp04,Tropp04relax}.

\subsubsection{Subspace Clustering with BP}
In \citep{vidaldisjoint}, the authors develop the following sufficient condition for EFS to occur for BP from a union of disjoint subspaces,
\begin{equation}
\label{eq:vidal}
\max_{j \ne i} ~ \cos(\theta_{ij}^{\ast}) ~ < ~\max_{ \widetilde{Y}_i \in \mathbb{W}_i} \frac{\sigma_{\min}(\widetilde{Y}_i)}{\sqrt{k_i}},
\end{equation}
where $\mathbb{W}_i$ is the set of all full rank sub-matrices $\widetilde{Y}_i \in \R^{n \times k_i}$ of the data matrix $Y_i \in \R^{n  \times k_i}$ and $\sigma_{\min}( \widetilde{Y}_i )$ is the minimum singular value of the sub-matrix $\widetilde{Y}_i$. Since we assume that all of the data points have been normalized,  $\sigma_{\min}( \widetilde{Y}_i ) \le 1$; thus, the best case result that can be obtained is that the minimum principal angle, $\cos(\theta_{ij}^{\ast})  < 1/\sqrt{k_i}$. This suggests that the minimum principal angle of the union must go to zero, i.e., the union must consist of orthogonal subspaces, as the subspace dimension increases. 

In contrast to the condition in (\ref{eq:vidal}), the conditions we provide in Thm. \ref{theo:EFS} and Cor.\ \ref{coro:EFSdisjoint} do not depend on the subspace dimension. Rather, we require that there are enough points in each subspace to achieve a sufficiently small covering; in which case, EFS can be guaranteed for subspaces of any dimension. 


In \citep{Candes12}, the authors develop the following sufficient condition for EFS from unions of intersecting subspaces with BP:
\begin{equation}
\label{eq:candesbound}
\mu_v(\set_i) = \max_{y \in \set_{-i}} \| {V_{(i)}}^T y \|_{\infty}  < r(\set_i),
\end{equation}
where the matrix $V_{(i)} \in \R^{d_i \times n}$ contains the dual directions (the dual vectors for each point in $\set_i$ embedded in $\R^n$) in its columns,{\footnote{See Def. 2.2 in \cite{Candes12} for a formal definition of the dual directions and insight into the geometry underlying their guarantees for EFS via BP.}} and $r(\set_i)$ is the inradius as defined in (\ref{eq:inradius}). In words, (\ref{eq:candesbound}) requires that the maximum coherence between any point in $\set_{-i}$ and the dual directions contained in $V_{(i)}$ be less than the inradius of the points in $\set_i$. 

To link the result in (\ref{eq:candesbound}) to our guarantee for OMP in Thm. \ref{theo:EFS}, we observe that while (\ref{eq:candesbound}) requires that $\mu_v (\set_i)$ be less than the inradius, Thm. \ref{theo:EFS} requires that the mutual coherence $\mu_c(\set_i)$ be less than the inradius minus an additional term that depends on the covering radius. For an arbitrary set of points that live on a union of subspaces, the precise relationship between the two coherence parameters $\mu_c(\set_i)$ (coherence between two points in different subspace clusters) and $\mu_v(\set_i)$ (coherence between a point in a subspace cluster and the dual directions of points in a different subspace cluster) is not straightforward; however, when the points in each subspace cluster are distributed uniformly and at random along each subspace, the dual directions will also be distributed uniformly along each subspace.\footnote{{This approximation is based upon personal correspondence with M. Soltankotabi, one of the authors that developed the results for EFS with BP in \cite{Candes12}.}} In this case, $\mu_v(\set_i)$ will be roughly equivalent to the mutual coherence $\mu_c(\set_i)$.




This simplification reveals the connection between the result in (\ref{eq:candesbound}) for BP and the condition in Thm.\ \ref{theo:EFS} for OMP. In particular, when $\mu_v(\set_i) \approx \mu_c(\set_i)$, our result for OMP requires that the mutual coherence is smaller than the inradius minus an additional term that is linear in the covering diameter $\epsilon$. For this reason, our result in Thm. \ref{theo:EFS} is more restrictive than the result in (\ref{eq:candesbound}). The gap between the two bounds shrinks to zero only when the minimum principal angle $\theta_{ij}^{\ast} \to \pi/2$ or when the covering diameter $\epsilon \to 0$. 

In our empirical studies, we find that when BPDN is tuned to an appropriate value of the noise parameter $\kappa$, BPDN does in fact provide higher rates of EFS than OMP.{\footnote{While BPDN provides higher rates of EFS than OMP when we employ a homotopy approach to find an optimal value of the noise parameter $\kappa$, for a wide range of values of $\kappa$, BPDN and OMP provide comparable rates of EFS.}} This suggests that the theoretical gap between the two approaches might not be an artifact of our current analysis; rather, there might exist an intrinsic gap between the performance of each method with respect to EFS. Nonetheless, an interesting finding from our empirical study in Section \ref{sec:facesubs}, is that despite the fact that BPDN provides better rates of EFS than OMP, OMP often provides better clustering results than BPDN. For these reasons, we maintain that OMP offers a powerful low-complexity alternative to $\ell_1$-minimization approaches for feature selection.








\subsubsection{Exact Recovery Conditions for Sparse Recovery}
To provide further intuition about EFS in endogenous sparse recovery, we will compare the geometry underlying the EFS condition with the geometry of the exact recovery condition (ERC) for sparse signal recovery methods provided in \citep{Tropp04,Tropp04relax}. 

To guarantee exact support recovery for a signal $y \in \R^n$ which has been synthesized from a linear combination of atoms from the sub-matrix $\dct_{\Lambda} \in \R^{n \times k}$, we must ensure that we can recover an approximation of $y$ that consists solely of atoms from $\dct_{\Lambda}$. Let $\{ \varphi_i \}_{i \notin \Lambda}$ denote the set of atoms in $\dct$ that are not indexed by the set $\Lambda$. The {\em exact recovery condition} (ERC) provided below in Thm.\ref{theo:ERC} is sufficient to guarantee that we obtain exact support recovery for both BP and OMP. 
\begin{THEO}{\bf \citep{Tropp04}} \label{theo:ERC} For any signal supported over the sub-dictionary $\dct_{\Lambda}$, exact support recovery is guaranteed for both OMP and BP if
\begin{equation}
\label{eq:erc} 
{{\rm ERC}}(\Lambda) = \max_{i \notin \Lambda} \|{{\dct}_{\Lambda}}^{\dagger} {\atom}_i\|_1 < 1.
\end{equation}
\end{THEO} 
A geometric interpretation of the ERC is that it provides a measure of how far a projected atom $\varphi_i$ outside of the set $\Lambda$ lies from the antipodal convex hull of the atoms in $\Lambda$. When a projected atom lies outside of the antipodal convex hull formed by the set of points in the sub-dictionary $\dct_{\Lambda}$, then the ERC condition is violated and support recovery is not guaranteed. 
For this reason, the ERC requires that the maximum coherence between the atoms in $\dct$ is sufficiently low or that $\dct$ is {\em incoherent}. 

While the ERC condition requires a {\em global incoherence} property on all of the columns of $\dct$, we can interpret EFS as requiring a {\em local incoherence} property. In particular, the EFS condition requires that the projection of atoms in an incorrect subspace cluster $\set_{-i}$ onto $\sub_i$ must be incoherent with any deep holes in $\set_i$ along $\sub_i$. In addition, we need that the points within a subspace cluster are {\em coherent} in order to produce a small covering radius.

\section{EFS for Uniformly Bounded Unions of Subspaces}
\label{sec:efsbounded}
In this section, we study the connection between EFS and the higher-order principal angles (beyond the minimum angle) between pairs of intersecting subspaces.


\subsection{Subspace Distances}
To characterize the ``distance'' between pairs of subspaces in the ensemble, the {\em principal angles} between subspaces will prove useful. As we saw in the previous section, the first principal angle $\theta_0$ between subspaces $\sub_1$ and $\sub_2$ of dimension $k_1$ and $k_2$ is defined as the smallest angle between a pair of unit vectors $(u_1,v_1)$ drawn from $\sub_1 \times \sub_2$. The vector pair $(u_1^*, v_1^*)$ that attains this minimum is referred to as the first set of principal vectors. The second principal angle $\theta_1$ is defined much like the first, except that the second set of principal vectors that define the second principal angle are required to be orthogonal to the first set of principal vectors $(u_1^*, v_1^*)$. The remaining principal angles are defined recursively in this way. The sequence of $k = \min(k_1, k_2)$ principal angles, $\theta_0 \le \theta_1 \le \cdots \le \theta_{k-1}$, is non-decreasing and all of the principal angles lie between $[0, \pi/2]$.  

The definition above provides insight into what the principal angles/vectors tell us about the geometry underlying a pair of subspaces; in practice, however, the principal angles are not computed in this recursive manner. Rather, a computationally efficient way to compute the principal angles between two subspaces $\sub_i$ and $\sub_j$ is to first compute the singular values of the matrix $G = \dct_i^T\dct_j$, where $\dct_i \in \R^{n \times k_i}$ is an ONB that spans subspace $\sub_i$. Let $G = U \Sigma V^T$ denote the SVD of $G$ and let $\sigma_{ij} \in [0,1]^k$ denote the singular values of $G$, where $k = \min ( k_i, k_j )$ is the minimum dimension of the two subspaces. The $m^{\rm th}$ smallest principal angle $\theta_{ij}(m)$ is related to the $m^{\rm th}$ largest entry of $\sigma_{ij}$ via the following relationship, $ \cos( \theta_{ij}(m)) = \sigma_{ij}(m)$.  For our subsequent discussion, we will refer to the singular values of $G$ as the {\em cross-spectra} of the subspace pair ($\sub_i,\sub_j$). 
 
A pair of subspaces is said to be {\em disjoint} if the minimum principal angle is greater than zero. Non-disjoint or intersecting subspaces are defined as subspaces with minimum principal angle equal to zero. The dimension of the intersection between two subspaces is equivalent to the number of principal angles equal to zero or equivalently, the number of entries of the cross-spectra that are equal to one. The {\em overlap} between two subspaces is defined as the ${\rm rank}(G)$ or equivalently, $q = \| \sigma_{ij} \|_0$, where $q \ge {\rm dim}(\sub_i \cap \sub_j$).

\subsection{Conditions for EFS from Bounded Unions} 
\label{sec:boundedunion}
The sufficient conditions for EFS in Thm. \ref{theo:EFS} and Cor.\ \ref{coro:EFSdisjoint} reveal an interesting relationship between the covering radius and the minimum principal angle between pairs of subspaces in the ensemble. However, we have yet to reveal any dependence between EFS and higher-order principal angles. To make this connection more apparent, we will make additional assumptions about the distribution of points in the ensemble, namely that the dataset produces an {\em uniformly bounded union} of subspaces relative to the principal vectors supporting pairs of subspaces in the ensemble. 


Let $Y = [Y_i~ Y_j]$ denote a collection of unit-norm data points, where $Y_i$ and $Y_j$ contain the points in subspaces $\sub_i$ and $\sub_j$, respectively. Let $G = \dct_i^T \dct_j = U \Sigma V^T$ denote the SVD of $G$, where ${\rm rank}(G) = q$ and let $\widetilde{U} = \dct_i U_q$ denote the set of left principal vectors of $G$ that are associated with the $q$ nonzero singular values in $\Sigma$. Similarly, let $\widetilde{V} = \dct_j V_q$ denote the set of right principal vectors of $G$ that are associated with the nonzero singular values in $\Sigma$. When the points in each subspace are incoherent with the principal vectors in the columns of $\widetilde{U}$ and $\widetilde{V}$, we say that the ensemble $Y$ is an {\em uniformly bounded union of subspaces}. Formally, we require the following incoherence property holds:
\begin{equation}\label{eq:boundunion} \left( \| Y_i^T \widetilde{U} \|_{\infty}, \| Y_j^T \widetilde{V} \|_{\infty} \right) \le \gamma, \end{equation}
where $\| \cdot \|_{\infty}$ is the entry-wise maximum and $\gamma \in (0,1]$. This property requires that the inner products between the points in a subspace and the set of principal vectors that span non-orthogonal directions between a pair of subspaces is bounded by a fixed constant. 





When the points in each subspace are distributed such that (\ref{eq:boundunion}) holds, we can rewrite the mutual coherence between any two points from different subspaces to reveal its dependence on higher-order principal angles. In particular, we show (in Section \ref{sec:proofboundedsubs}) that the coherence between the residual $s$ used in Alg.\ \ref{alg:omp} to select the next point to be included in the representation of a point $y \in \set_i$, and a point in $\set_j$ is upper bounded by
\begin{equation}
\label{eqn:tackle2}
\max_{y \in \set_j}\frac{ | \langle s, y \rangle | } { \| s\|_2 }  \le \gamma \| \sigma_{ij} \|_1,
\end{equation}
where $\gamma$ is the bounding constant of the data $Y$ and $\| \sigma_{ij} \|_1$ is the $\ell_1$-norm of the cross-spectra or equivalently, the trace norm of $G$. Using the bound in (\ref{eqn:tackle2}), we arrive at the following sufficient condition for EFS from uniformly bounded unions of subspaces.
\begin{THEO} \label{theo:boundedvecs} Let $Y$ be a uniformly bounded union of subspaces as defined in (\ref{eq:boundunion}), where $q = {\rm rank}(G)$, and $\gamma < \sqrt{1/q}$. Let $\sigma_{ij}$ denote the cross-spectra of the subspaces $\sub_i$ and $\sub_j$. A sufficient condition for EFS to occur for all of the points in $\set_i$ is that the covering diameter 
$$ \epsilon~ <~  \min_{j \ne i}~ \sqrt{ 1 - \gamma^2 \| \sigma_{ij} \|_1^2 }.$$
\end{THEO}
This condition requires that both the covering diameter of each subspace and the bounding constant of the union be sufficiently small in order to guarantee EFS. One way to guarantee that the ensemble has a small bounding constant is to constrain the total amount of energy that points in $\set_j$ have in the $q$-dimensional subspace spanned by the principal vectors in $\widetilde{V}$.

Our analysis for bounded unions assumes that the nonzero entries of the cross-spectra are equal, and thus each pair of supporting principal vectors in $\widetilde{V}$ are equally important in determining whether points in $\set_i$ will admit EFS. However, this assumption is not true in general. When the union is supported by principal vectors with non-uniform principal angles, our analysis suggests that a weaker form of incoherence is required. Instead of requiring incoherence with all principal vectors, the data must be sufficiently incoherent with the principal vectors that correspond to small principal angles (or large values of the cross-spectra). This means that as long as points are not concentrated along the principal directions with small principal angles (i.e., intersections), then EFS can be guaranteed, even when subspaces exhibit non-trivial intersections. To test this prediction, we will study a {\em bounded energy model} for the unions of subspaces in Section \ref{sec:phasetrans}. We show that when the dataset is sparsely sampled (larger covering radius), reducing the amount of energy that points contain in the intersections between two subspaces, increases the probability that points admit EFS dramatically. 

Finally, our analysis of bounded unions suggests that the decay of the cross-spectra is likely to play an important role in determining whether points will admit EFS or not. To test this hypothesis, we will study the role that the structure of the cross-spectra plays in EFS in Section \ref{sec:ell0vsNN}. 

\section{Experimental Results}
\label{sec:results}
In our theoretical analysis of EFS in Sections \ref{sec:efs_union} and \ref{sec:efsbounded}, we revealed an intimate connection between the covering radius of subspaces and the principal angles between pairs of subspaces in the ensemble. In this section, we will conduct an empirical study to explore these connections further. In particular, we will study the probability of EFS as we vary the covering radius as well as the dimension of the intersection and/or overlap between subspaces. In addition, we will study the role that the structure of the cross-spectra and the amount of energy that points have in subspace intersections, have on EFS.

\subsection{Generative Model for Synthetic Data}
In order to study EFS for unions of subspaces with varied cross-spectra, we will generate synthetic data from unions of overlapping {\em block-sparse signals}. 

\subsubsection{Constructing Sub-dictionaries}
\label{sec:crosspec}
We construct a pair of sub-dictionaries as follows: Take two subsets $\Omega_1$ and $\Omega_2$ of $k$ atoms from a dictionary $D$ containing $M$ atoms $\{ d_m \}_{m=1}^M$ in its columns, where $d_m \in \R^{n}$ and $|\Omega_1| = |\Omega_2| = k$. Let $\Psi \in \R^{n \times k}$ denote the subset of atoms indexed by ${\Omega_1},$ and let $\Phi \in \R^{n \times k}$ denote the subset of atoms indexed by ${\Omega_2}$. Our goal is to select $\Psi$ and $\Phi$ such that $G = \Psi^T \Phi$ is diagonal, i.e., $\langle \psi_i , \phi_j \rangle = 0,$ if $i \ne j$, where $\psi_i$ is the $i^{\rm th}$ element in $\Psi$ and $\phi_j$ is the $j^{\rm th}$ element of $\Phi$. In this case, the cross-spectra is defined as $\sigma = {\rm diag}(G)$, where $\sigma \in [0,1]^k$. For each union, we fix the ``overlap'' $q$ or the rank of $G = \Psi^T \Phi$ to a constant between zero (orthogonal subspaces) and $k$ (maximal overlap).

To generate a pair of $k$-dimensional subspaces with a $q$-dimensional overlap, we can pair the elements from $\Psi$ and $\Phi$ such that the $i^{\rm th}$ entry of the cross-spectra equals 
\begin{equation*}
\label{eq:sigma}
\sigma(i) = 
\begin{cases} | \langle \psi_i , \phi_i \rangle | \quad & \text{if $1\le i \le q$,}
\\
0 \quad &\text{if $i = q+1\le i \le k.$}
\end{cases}
\end{equation*}

We can leverage the banded structure of shift-invariant dictionaries, e.g., dictionary matrices with localized Toeplitz structure, to generate subspaces with structured cross-spectra as follows.{\footnote{While shift-invariant dictionaries appear in a wide range of applications of sparse recovery \citep{mailhe2008,dyerlva}, we introduce the idea of using shift-invariant dictionaries to create structured unions of subspaces for the first time here.} First, we fix a set of $k$ incoherent (orthogonal) atoms from our shift-invariant dictionary, which we place in the columns of $\Psi$. Now, holding $\Psi$ fixed, we set the $i^{\rm th}$ atom $\phi_i$ of the second sub-dictionary $\Phi$ to be a shifted version of the $i^{\rm th}$ atom $\psi_i$ of the dictionary $\Psi$. To be precise, if we set $\psi_i = d_m$, where $d_m$ is the $m^{\rm th}$ atom in our shift-invariant dictionary, then we will set $\phi_i = d_{m+\Delta}$ for a particular shift $\Delta$. By varying the shift $\Delta$, we can control the coherence between $\psi_i$ and $\varphi_i$. In Fig.\ \ref{fig:shiftatoms}, we show an example of one such construction for $k = q = 5$.  Since $\sigma \in (0,1]^k$, the worst-case pair of subspaces with overlap equal to $q$ is obtained when we pair $q$ identical atoms with $k-q$ orthogonal atoms. In this case, the cross-spectra attains its maximum over its entire support and equals zero otherwise. For such unions, the overlap $q$ equals the dimension of the intersection between the subspaces. We will refer to this class of block-sparse signals as {\em orthoblock sparse signals}. 

\label{sec:blksparse}
\begin{figure}[t!]
\vskip 0.2in
\begin{center}
\centerline{\includegraphics[width=6in]{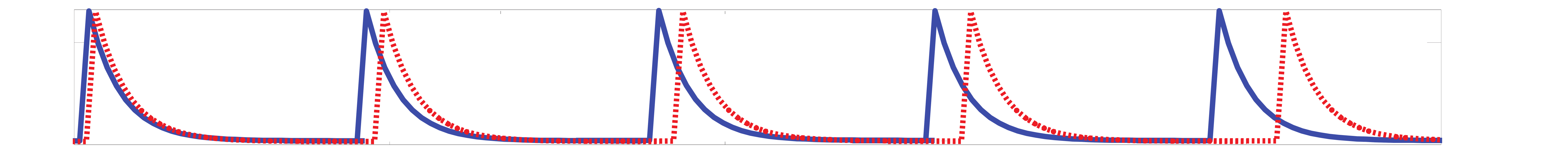}}
\caption{ \label{fig:shiftatoms} {\em Generating unions of subspaces from shift-invariant dictionaries.}~An example of a collection of two sub-dictionaries of five atoms, where each of the atoms have a non-zero inner product with one other atom. This choice of sub-dictionaries produces a union of disjoint subspaces, where the overlap ratio $\delta = q/k = 1$.}
\end{center}
\vskip -0.2in
\end{figure}

\subsubsection{Coefficient Synthesis}
\label{sec:coeffs}
To synthesize a point that lives in the span of the sub-dictionary $\Psi \in \R^{n \times k}$, we combine the elements $\{ \psi_1, \dots, \psi_k \}$ and subspace coefficients $\{ \alpha(1), \dots, \alpha(k) \}$ linearly to form
$$y_i =  \sum_{j=1}^k \psi_j \alpha(j),$$ 
where $\alpha(j)$ is the subspace coefficient associated with the $j^{\rm th}$ column in $\Psi$. Without loss of generality, we will assume that the elements in $\Psi$ are sorted such that the values of the cross-spectra are monotonically decreasing. Let $y_i^c = \sum_{j =1}^q \psi_j \alpha_i(j)$ be the ``common component'' of $y_i$ that lies in the space spanned by the principal directions between the pair of subspaces that correspond to non-orthogonal principal angles between $(\Phi,\Psi)$ and let $y_i^d = \sum_{j =q+1}^k \psi_j \alpha(j)$ denote the ``disjoint component'' of $y_i$ that lies in the space orthogonal to the space spanned by the first $q$ principal directions.

For our  experiments, we consider points drawn from one of the two following coefficient distributions, which we will refer to as {\emph{(M1)}} and {\emph{(M2)}} respectively.
\begin{itemize}
\item{\em (M1) Uniformly Distributed on the Sphere: }~Generate subspace coefficients according to a standard normal distribution and map the point to the unit sphere
 $$y_i =  \frac{ \sum_j \psi_j \alpha(j) }{ \|  \sum_j \psi_j \alpha(j)  \|_2 }, \quad {\rm where} ~~~ \alpha(j) \sim \mathcal{N}(0,1).$$
\item{\em (M2) Bounded Energy Model: }~Generate subspace coefficients according to {\emph{(M1)}} and rescale each coefficient in order to bound the energy in the common component
$$y_i =  \frac{\tau  y_i^c} { \| y_i^c \|_2 } +  \frac{ (1-  \tau)  y_i^d} { \| y_i^d \|_2 }.$$
\end{itemize}
By simply restricting the total energy that each point has in its common component, the bounded energy model {\emph{(M2)}} can be used to produce ensembles with small bounding constant to test the predictions in Thm.\ \ref{theo:boundedvecs}.

\subsection{Phase Transitions for OMP}
\label{sec:phasetrans}
The goal of our first experiment is to study the probability of EFS---the probability that a point in the ensemble admits exact features---as we vary both the number and distribution of points in each subspace as well as the dimension of the intersection between subspaces. For this set of experiments, we generate a union of orthoblock sparse signals, where the overlap equals the dimension of the intersection.

\begin{figure}[t!]
\vskip 0.2in
\begin{center}
\centerline{\includegraphics[width= 4.5in]{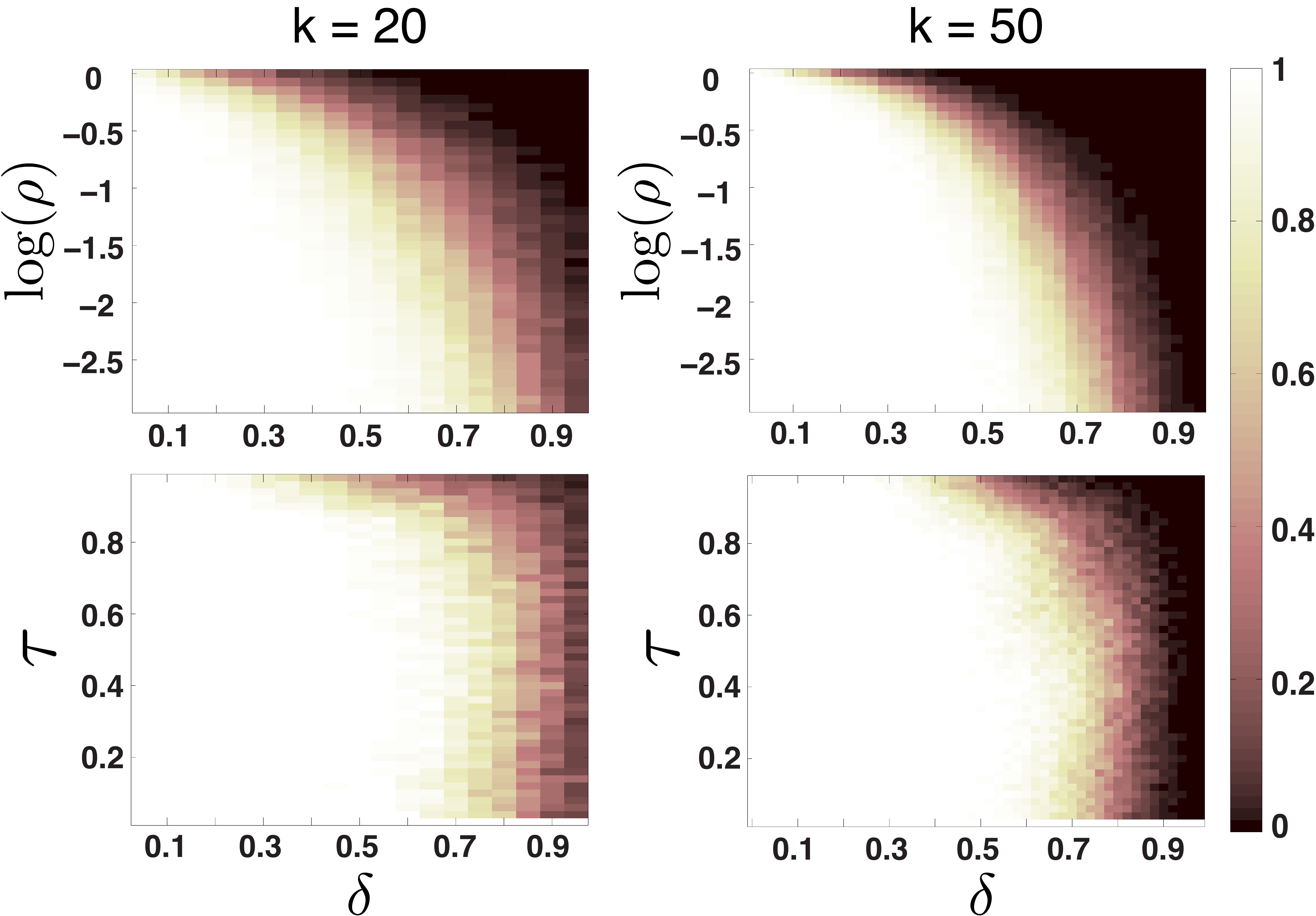}}
\caption[ {\em Probability of EFS for different sampling conditions}]{ \label{fig:oversamp} {\em Probability of EFS for different coefficient distributions.}~The probability of EFS for a union of two subspaces of dimension $k = 20$ (left column) and $k = 50$ (right column). The probability of EFS is displayed as a function of the overlap ratio $\delta \in [0,1)$ and the logarithm of the oversampling ratio $\log(\rho)$ (top row) and the mutual energy $\tau = \| y_c \|_2$ (bottom row) .}
\end{center}
\vspace{-4mm}
\end{figure}

Along the top row of Fig.\ \ref{fig:oversamp}, we display the probability of EFS for orthoblock sparse signals generated according to the coefficient model {\emph{(M1)}}: the probability of EFS is computed as we vary the {\em overlap ratio} $\delta = q/k \in [0,1]$ in conjunction with the {\em oversampling ratio} $\rho = k/d \in [0,1]$, where $q = {\rm rank}(\dct_1^T \dct_2)$ equals the dimension of the intersection between the subspaces, and $d$ is the number of points per subspace. Along the bottom row of Fig.\ \ref{fig:oversamp},  we display the probability of EFS for orthoblock sparse signals generated according to the coefficient model {\emph{(M2)}}: the probability of EFS is computed as we vary the overlap ratio $\delta$ and the amount of energy $\tau \in [0,1)$ each point has within its common component. For these experiments, the subspace dimension is set to $k = 20$ (left) and $k = 50$ (right). To see the phase boundary that arises when we approach critical sampling (i.e., $\rho \approx 1$), we display our results in terms of the logarithm of the oversampling ratio. For these experiments, the results are averaged over $500$ trials.

As our theory predicts, the oversampling ratio has a strong impact on the degree of overlap between subspaces that can be tolerated before EFS no longer occurs. In particular, as the number of points in each subspace increases (covering radius decreases), the probability of EFS obeys a second-order phase transition, i.e., there is a graceful degradation in the probability of EFS as the dimension of the intersection increases. When the pair of subspaces are densely sampled, the phase boundary is shifted all the way to $\delta = 0.7$, where$70\%$  of the dimensions of each subspace intersect. This is due to the fact that as each subspace is sampled more densely, the covering radius becomes sufficiently small to ensure that even when the overlap between planes is high, EFS still occurs with high probability. In contrast, when the subspaces are critically sampled, i.e., the number of points per subspace $d \approx k$, only a small amount of overlap can be tolerated, where $\delta < 0.1$. In addition to shifting the phase boundary, as the oversampling ratio increases, the width of the transition region (where the probability of EFS goes from zero to one) also increases.

Along the bottom row of Fig.\ \ref{fig:oversamp}, we study the impact of the bounding constant on EFS, as discussed in Section \ref{sec:boundedunion}. In this experiment, we fix the oversampling ratio to $\rho = 0.1$ and vary the common energy $\tau$ in conjunction with the overlap ratio $\delta$. By reducing the bounding constant of the union, the phase boundary for the uniformly distributed data from model {\emph{(M1)}} is shifted from $\delta = 0.45$ to $\delta = 0.7$ for both $k = 20$ and $k = 50$. This result confirms our predictions in the discussion of Thm.\ \ref{theo:boundedvecs} that by reducing the amount of energy that points have in their subspace intersections EFS will occur for higher degrees of overlap. Another interesting finding of this experiment is that, once $\tau$ reaches a threshold, the phase boundary remains constant and further reducing the bounding constant has no impact on the phase transitions for EFS. 



\subsection{Comparison of OMP and NN}
\label{sec:ell0vsNN}
In this section, we compare the probability of EFS for feature selection with OMP and nearest neighbors (NN). First, we compare the performance of both feature selection methods for unions with different cross-spectra. Second, we compare the phase transitions for unions of orthoblock sparse signals as we vary the overlap and oversampling ratio.
\begin{figure}[t!]
\vskip 0.2in
\begin{center}
\centerline{\includegraphics[width=5.5in]{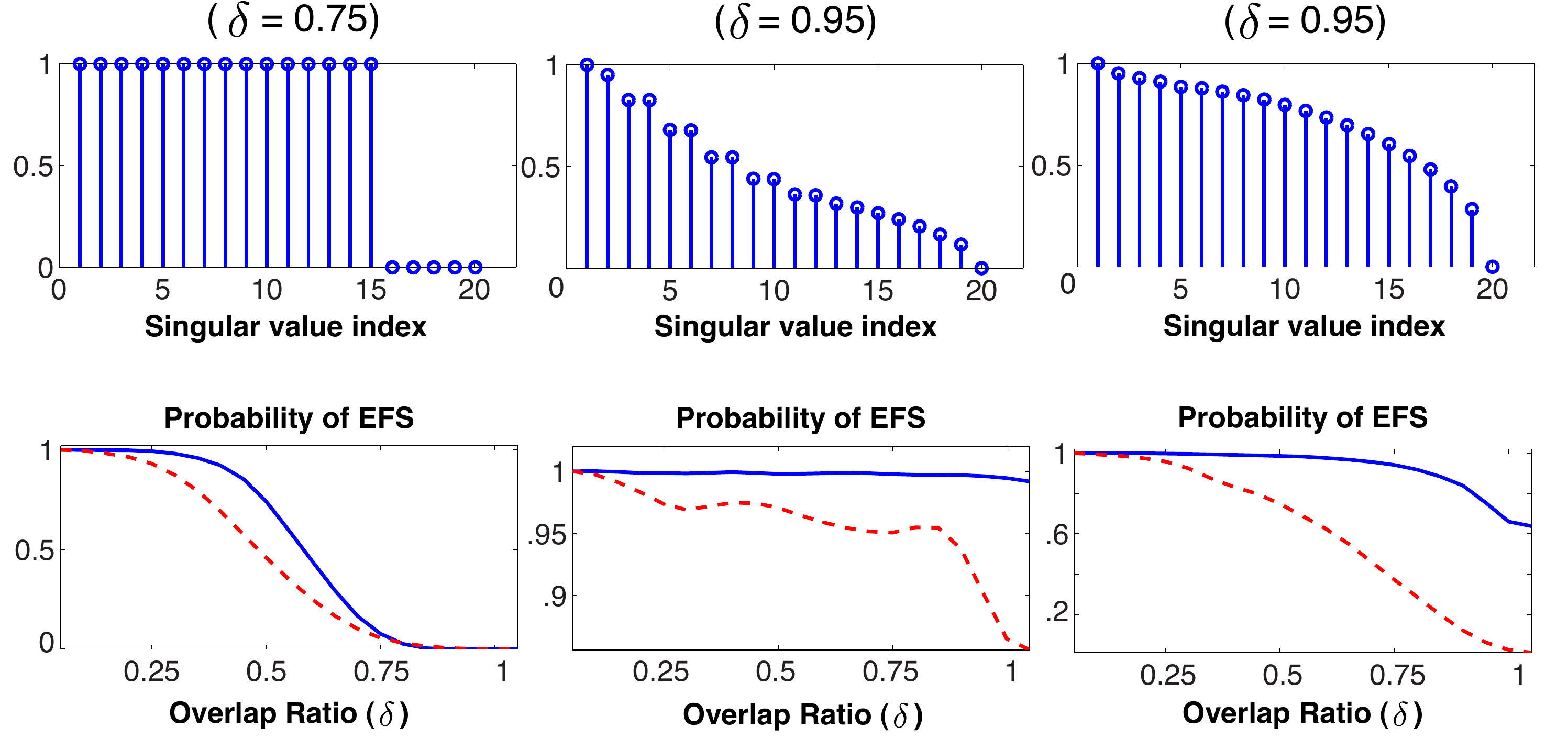}}
\caption[{\em Probability of EFS for unions with structured cross-spectra}]{ \label{fig:blksparse}  {\em Probability of EFS for unions with structured cross-spectra.} Along the top row, we show the cross-spectra for different unions of block-sparse signals. Along the bottom row, we show the probability of EFS as we vary the overlap ratio $\delta \in [0,1]$ for OMP (solid) and NN (dash).}
\end{center}
\vskip -0.2in
\end{figure}

For our experiments, we generate pairs of subspaces with structured cross-spectra as described in Section \ref{sec:crosspec}. The cross-spectra arising from three different unions of block-sparse signals are displayed along the top row of Fig.\ \ref{fig:blksparse}. On the left, we show the cross-spectra for a union of orthoblock sparse signals with overlap ratio $\delta = 0.75$, where $q=15$ and $k = 20$. The cross-spectra obtained by pairing shifted Lorentzian and exponential atoms are displayed in the middle and right columns, respectively. Along the bottom row of Fig.\ \ref{fig:blksparse}, we show the probability of EFS for OMP and NN for each of these three subspace unions as we vary the overlap $q$. To do this, we generate subspaces by setting their cross-spectra equal to the first $q$ entries equal to the cross-spectra in Fig.\ \ref{fig:blksparse} and setting the remaining $k-q$ entries of the cross-spectra equal to zero. Each subspace cluster is generated by sampling $d = 100$ points from each subspace according to the coefficient model {\emph{(M1)}}.



\begin{figure}[t!]
\vskip 0.2in
\begin{center}
\centerline{\includegraphics[width=4.2in]{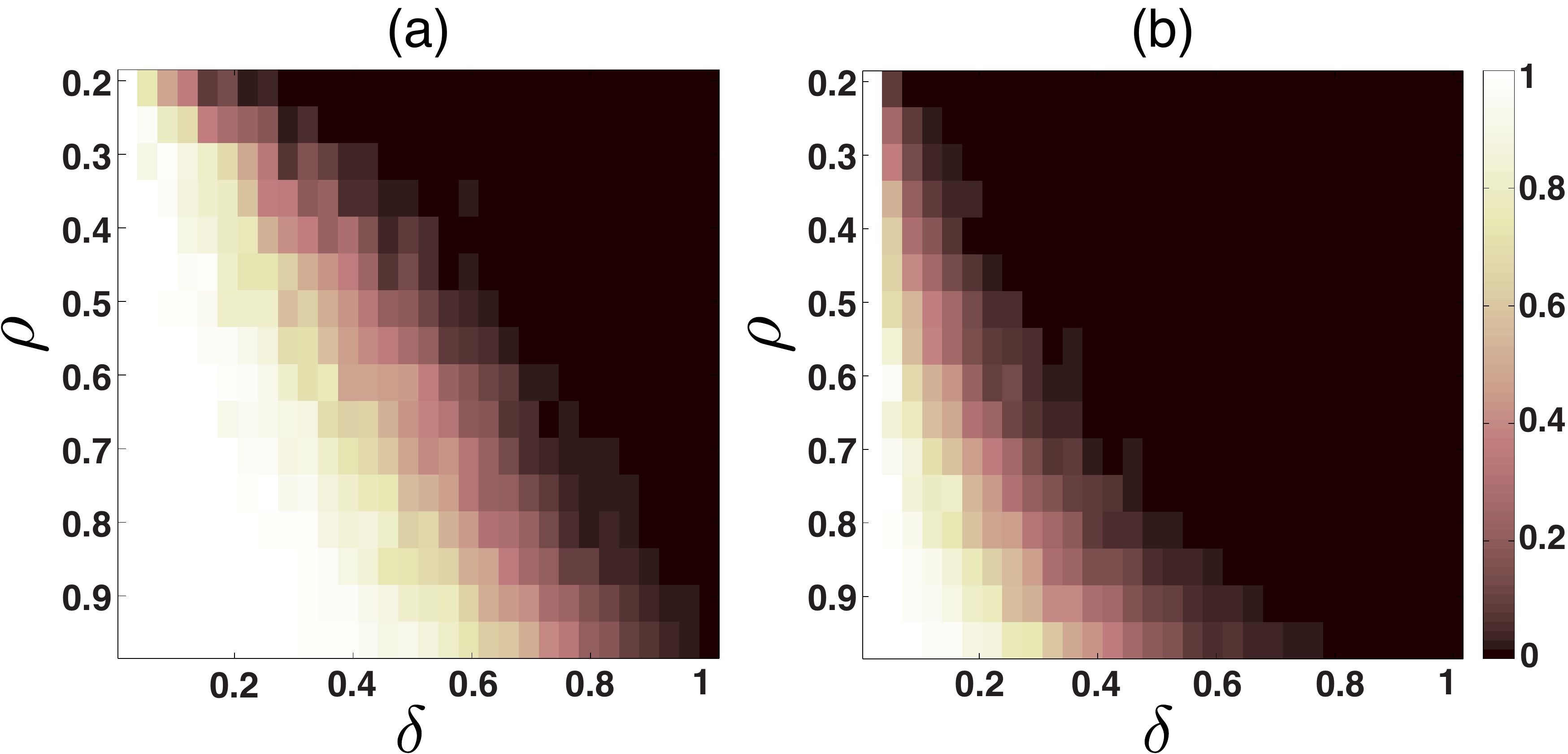}}
\caption[\em Phase transitions for sparse recovery and nearest neighbor graphs.]{\label{fig:phasetransknn}  {\em Phase transitions for OMP and NN.} The probability of EFS for orthoblock sparse signals for OMP (a) and NN (b) feature sets as a function of the oversampling ratio $\rho = k/d$ and the overlap ratio $\delta = q/k$, where $k = 20$.}
\end{center}
\vskip -0.3in
\end{figure}

This study provides a number of interesting insights into the role that higher-order principal angles between subspaces play in feature selection for both sparse recovery methods and NN. First, we observe that the gap between the probability of EFS for OMP and NN is markedly different for each of the three unions. In the first union of orthoblock sparse signals, the probability of EFS for OMP lies strictly above that obtained for the NN method, but the gap between the performance of both methods is relatively small. In the second union, both methods maintain a high probability of EFS, with OMP admitting nearly perfect feature sets even when the overlap ratio is maximal. In the third union, we observe that the gap between EFS for OMP and NN is most pronounced. In this case, the probability of EFS for NN sets decreases to $0.1$, while OMP admits a very high probability of EFS, even when the overlap ratio is maximal. in summary, we observe that when data is distributed uniformly with respect to all of the principal directions between a pair of subspaces and the cross-spectra is sub-linear, then EFS may be guaranteed with high probability for all points in the set provided the sampling density is sufficiently high. This is in agreement with the discussion of EFS bounded unions in Section \ref{sec:boundedunion}. Moreover, these results further support our claims that in order to truly understand and predict the behavior of endogenous sparse recovery from unions of subspaces, we require a description that relies on the entire cross-spectra.
\vspace{2mm}
 
In Fig.\ \ref{fig:phasetransknn}, we display the probability of EFS for OMP (left) and sets of NN (right) as we vary the overlap and the oversampling ratio. For this experiment, we consider unions of orthoblock sparse signals living on subspaces of dimension $k = 50$ and vary $\rho \in [0.2, 0.96]$ and $\delta \in [1/k, 1]$. An interesting result of this study is that there are regimes where the probability of EFS equals zero for NN but occurs for OMP with a non-trivial probability. In particular, we observe that when the sampling of each subspace is sparse (the oversampling ratio is low), the gap between OMP and NN increases and OMP significantly outperforms NN in terms of their probability of EFS. Our study of EFS for structured cross-spectra suggests that the gap between NN and OMP should be even more pronounced for cross-spectra with superlinear decay.

\subsection{Face Illumination Subspaces}
\label{sec:facesubs}
In this section, we compare the performance of sparse recovery methods, i.e., BP and OMP, with NN  for clustering unions of {\em illumination subspaces} arising from a collection of images of faces under different lighting conditions. By fixing the camera center and position of the persons face and capturing multiple images under different lighting conditions, the resulting images can be well-approximated by a $5$-dimensional subspace \citep{facesubs}. 

In Fig.\ \ref{fig:faces}, we show three examples of the subspace affinity matrices obtained with NN, BP, and OMP for two different faces under $64$ different illumination conditions from the Yale Database B \citep{yaleb}, where each image has been subsampled to $48 \times 42$ pixels, with $n = 2016$. In all of the examples, the data is sorted such that the images for each face are placed in a contiguous block. 

To generate the NN affinity matrices in the left column of Fig.\ \ref{fig:faces}, we compute the absolute normalized inner products between all points in the dataset and then threshold each row to select the $k = 5$ nearest neighbors to each point. To generate the OMP affinity matrices in the right column, we compute the sparse representations of each point in the dataset with Alg.\ \ref{alg:omp} for $k=5$ and stack the resulting coefficient vectors into the rows of a matrix $C$; the final subspace affinity $W$ is computed by symmetrizing the coefficient matrix, $W = |C| + |C^T|$. To generate the BP affinity matrices in the middle column, we solved the BP denoising (BPDN) problem in (\ref{eq:bpdn}) via a homotopy algorithm where we sweep over the noise parameter $\kappa$ and choose the smallest value of $\kappa$ that produces $k \le 5$ coefficients.{\footnote{We also studied another variant of BPDN where we solve OMP for $k=5$, compute the error of the resulting approximation, and then use this error as the noise parameter $\kappa$. However, this variant provided worse results than those reported in Table 1.} The resulting coefficient vectors are then stacked into the rows of a matrix $C$ and the final subspace affinity $W$ is computed by symmetrizing the coefficient matrix, $W = |C| + |C^T|$.

After computing the subspace affinity matrix for each of these three feature selection methods, we employ a spectral clustering approach which partitions the data based upon the eigenvector corresponding to the smallest nonzero eigenvalue of the graph Laplacian of the affinity matrix \citep{graphcuts,ng2002spectral}. For all three feature selection methods, we obtain the best clustering performance when we cluster the data based upon the graph Laplacian instead of the normalized graph Laplacian \citep{graphcuts}. In Table \ref{fig:facetable}, we display the percentage of points that resulted in EFS and the final clustering error for all pairs of $38 \choose 2$ subspaces in the Yale B database. Along the top row, we display the mean and median percentage of points that resulted in EFS for the full dataset (all $64$ illumination conditions), half of the dataset ($32$ illumination conditions selected at random in each trial), and a quarter of the dataset ($16$ illumination conditions selected at random in each trial).  Along the bottom row, we display the clustering error (percentage of points that were incorrectly classified) for all three methods.

While both sparse recovery methods (BPDN and OMP) admit EFS rates that are comparable to NN on the full dataset, we find that sparse recovery methods provide higher rates of EFS than NN when the sampling of each subspace is sparse, i.e., the half and quarter datasets. These results are also in agreement with our experiments on synthetic data. A surprising result is that OMP provides better clustering performance than BP on this particular dataset, even though OMP has lower rates of EFS.


\begin{table}[t!]
\vskip 0.2in
\begin{center}
\centerline{\includegraphics[width=\columnwidth]{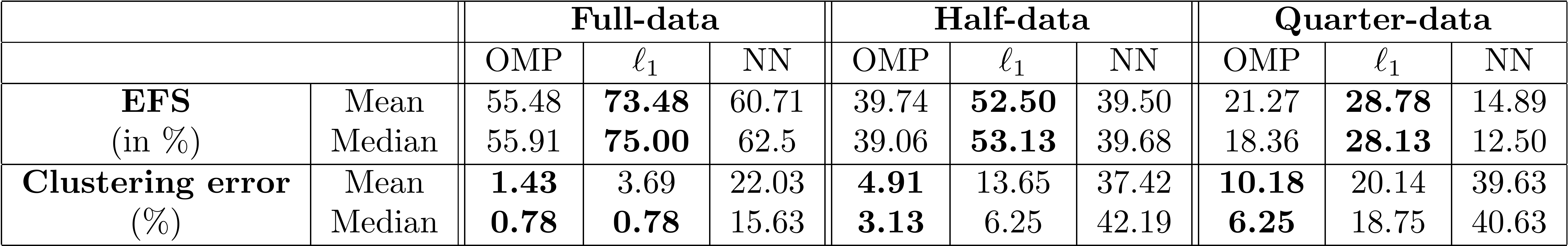}}
\caption[{\em Classification and EFS rates for illumination subspaces.}] {\label{fig:facetable} {\em Classification and EFS rates for illumination subspaces.} Shown are the aggregate results obtained over $38 \choose 2$ pairs of subspaces.}
\end{center}
\vskip -0.3in
\end{table}

\section{Discussion}
In this section, we provide insight into the implications of our results for different applications of sparse recovery and compressive sensing. Following this, we end with some open questions and directions for future research.

\subsection{``Data Driven'' Sparse Approximation}
The standard paradigm in signal processing and approximation theory is to compute a representation of a signal in a fixed and pre-specified basis or overcomplete dictionary. In most cases, the dictionaries used to form these representations are designed according to some mathematical desiderata. A more recent approach has been to learn a dictionary from a collection of data, such that the data admit a sparse representation with respect to the learned dictionary \citep{bruno97, ksvd}.

The applicability and utility of endogenous sparse recovery in subspace learning draws into question whether we can use endogenous sparse recovery for other tasks, including approximation and compression. The question that naturally arises is, ``do we design a dictionary, learn a dictionary, or use the data as a dictionary?'' Understanding the advantages and tradeoffs between each of these approaches is an interesting and open question.

\subsection{Learning Block-Sparse Signal Models}
Block-sparse signals and other structured sparse signals have received a great deal of attention over the past few years, especially in the context of compressive sensing from structured unions of subspaces \citep{unionsubs,Blu} and in model-based compressive sensing \citep{modelcs}. In all of these settings, the fact that a class or collection of signals admit structured support patterns is leveraged in order to obtain improved recovery of sparse signals in noise and in the presence of undersampling. 

To exploit such structure in sparse signals---especially in situations where the structure of signals or blocks of active atoms may be changing across different instances in time, space, etc.---the underlying subspaces that the signals occupy must be learned directly from the data. The methods that we have described for learning union of subspaces from ensembles of data can be utilized in the context of learning block sparse and other structured sparse signal models. The application of subspace clustering methods for this purpose is an interesting direction for future research.

\subsection{Beyond Coherence}
While the maximum and cumulative coherence \citep{Tropp04} provide measures of the uniqueness of sub-dictionaries that are necessary to guarantee exact signal recovery for sparse recovery methods, our current study suggests that examining the principal angles formed from pairs of sub-dictionaries could provide an even richer description of the geometric properties of a dictionary. Thus, a study of the principal angles formed by different subsets of atoms from a dictionary might provide new insights into the performance of sparse recovery methods with coherent dictionaries and for compressive sensing from structured matrices. In addition, our empirical results in Section \ref{sec:ell0vsNN} suggest that there might exist an intrinsic difference between sparse recovery from dictionaries that exhibit sublinear versus superlinear decay in their principal angles or cross-spectra. It would be interesting to explore whether these two ``classes'' of dictionaries exhibit different phase transitions for sparse recovery.


\subsection{Discriminative Dictionary Learning}
While dictionary learning was originally proposed for learning dictionaries that admit sparse representations of a collection of signals \citep{bruno97,ksvd}, dictionary learning has recently been employed for classification. To use learned dictionaries for classification, a dictionary is learned for each class of training signals and then a sparse representation of a test signal is formed with respect to each of the learned dictionaries. The idea is that the test signal will admit a more compact representation with respect to the dictionary that was learned from the class of signals that the test signal belongs to. 

Instead of learning these dictionaries independently of one another, {\em discriminative dictionary learning} \citep{mairal2008discriminative, ramirez2010}, aims to learn a collection of dictionaries $\{ \dct_1, \dct_2, \dots, \dct_p\}$  that are incoherent from one another. This is accomplished by minimizing either the spectral or Frobenius norm of the matrix product $\dct_i^T \dct_j$ between pairs of dictionaries. This same approach is utilized in \citep{Mailhe2012} to learn sensing matrices for CS that are incoherent with a learned dictionary.

There are a number of interesting connections between discriminative dictionary learning and our current study of EFS from collections of unions of subspaces. In particular, our study provides new insights into the role that the principal angles between two dictionaries tell us about our ability to separate classes of data based upon their sparse representations. Our study of EFS from unions with structured cross-spectra suggests that the decay of the cross-spectra between different data classes provides a powerful predictor of the performance of sparse recovery methods from data living on a union of low-dimensional subspaces. This suggests that in the context of discriminative dictionary learning, it might be more advantageous to reduce the $\ell_1$-norm of the cross-spectra rather than simply minimizing the maximum coherence and/or Frobenius norm between points in different subspaces as in \citep{mairal2008discriminative} and \citep{ramirez2010} respectively. To do this, each class of data must first be embedded within a subspace, a ONB is formed for each subspace, and then the $\ell_1$- norm of the cross-spectra must be minimized. An interesting question is how one might impose such a constraint in discriminative dictionary learning methods.

\subsection{Open Questions and Future Work}

While EFS provides a natural measure of how well a feature selection algorithm will perform for the task of subspace clustering, our empirical results suggest that EFS does not necessarily predict the performance of spectral clustering methods when applied to the resulting subspace affinity matrices. In particular, we find that while OMP obtains lower rates of EFS than BP on real-world data, OMP yields better clustering results on the same dataset.  Understanding where this difference in performance might arise from is an interesting direction for future research.

Another interesting finding of our empirical study is that the gap between the rates of EFS for sparse recovery methods and NN depends on the sampling density of each subspace. In particular, we found that for dense samplings of each subspace, the performance of NN is comparable to sparse recovery methods; however, when each subspace is more sparsely sampled, sparse recovery methods provide significant gains over NN methods. This result suggests that endogenous sparse recovery provides a powerful strategy for clustering when the sampling of subspace clusters is sparse. Analyzing the gap between sparse recovery methods and NN methods for feature selection is an interesting direction for future research.


Other directions for future research include: extending our deterministic analysis to random and semi-random settings such as those provided in \citep{Candes12} and studying the performance of OMP on noisy or corrupted data living on unions of subspaces. 

\section{Proofs}
\label{sec:proofs}

\subsection{Proof of Theorem \ref{theo:EFS}}
\label{sec:proofefstheo}
Our goal is to prove that, if (\ref{eq:efs}) holds, 
then it is sufficient to guarantee that EFS occurs for every point in $\set_k$ when OMP is used for feature selection. We will prove this  by induction. 

Consider the greedy selection step in OMP (see Alg.\ \ref{alg:omp}) for a point $y_i$ which belongs to the subspace cluster $\set_k$. Recall that at the $m^{th}$ step of OMP, the point that is maximally correlated with the signal residual will be selected to be included in the feature set $\Lambda$. The normalized residual at the $m^{\rm th}$ step is computed as
 \begin{equation}
 \label{eq:normresid}
s^{m} = \frac{(I  - P_{\Lambda}) y_i}{\| (I  - P_{\Lambda}) y_i \|_2},
\end{equation} 
where $P_{\Lambda} = Y_{\Lambda} Y_{\Lambda}^{\dagger} \in \R^{n \times n}$ is a projector onto the subspace spanned by the points in the current feature set $\Lambda$, where $|\Lambda| = m-1$.

To guarantee that we select a point from $\mathcal{S}_k$, we require that the following greedy selection criterion holds:
\begin{equation}
\label{eq:selectratio}
\max_{v \in \set_k}~~  |\langle s^{m} , v  \rangle |   >   \max_{v \notin \set_k} ~~  | \langle s^{m},  v \rangle |.
\end{equation}

We will prove that this selection criterion holds at each step of OMP by developing an upper bound on the RHS (the maximum inner product between the residual and a point outside of $\set_k$) and a lower bound on the LHS (the minimum inner product between the residual and a point in $\set_k$).

First, we will develop the upper bound on the RHS. In the first iteration, the residual is set to the signal of interest ($y_i$). In this case, we can bound the RHS by the mutual coherence $\mu_c = \max_{i \ne j} \mu_c(\set_i,\set_j)$ across all other sets
$$
\max_{y_j \notin \set_k} ~~  | \langle y_i,  y_j \rangle | \le \mu_c.
$$
Now assume that at the $m^{\rm th}$ iteration we have selected points from the correct subspace cluster. This implies that our signal residual still lies within the span of $\set_k$, and thus we can write the residual $s^m = z + e$, where $z$ is the closest point to $s^m$ in $\set_k$ and $e$ is the remaining portion of the residual which also lies in $\sub_k$. Thus, we can bound the RHS as follows
\begin{align*}
\max_{y_j \notin \set_k} | \langle s^{m},  y_j \rangle |  &= \max_{y_j \notin \set_k} | \langle z + e,  y_j \rangle |\\
&\le \max_{y_j \notin \set_k} | \langle z,  y_j \rangle | + | \langle e,  y_j \rangle |\\
&\le \mu_c + \max_{y_j \notin \set_k} | \langle e,  y_j \rangle |\\
&\le \mu_c  + \cos(\theta_0) \| e \|_2 \| y_i \|_2.
\end{align*}
Using the fact that ${\rm cover}(\set_k) = \epsilon/2$, we can bound the $\ell_2$-norm of the vector $e$ as
\begin{align*}
\| e \|_2 &= \| s - z \|_2\\
&=\sqrt{  \| s \|_2^2 + \| z \|_2^2 - 2 | \langle s, z \rangle |}\\
&\le \sqrt{ 2 - 2 \sqrt{1 - (\epsilon/2)^2}}\\
&=\sqrt{ 2 - \sqrt{4 - \epsilon^2}}.
\end{align*}
Plugging this quantity into our expression for the RHS, we arrive at the following upper bound
$$\max_{y_j \notin \set_k} | \langle s^{m},  y_j \rangle | \le \mu_c  + \cos(\theta_0) \sqrt{ 2 - \sqrt{4 - \epsilon^2}} <  \mu_c  +   \cos(\theta_0) \frac{ \epsilon}{ \sqrt[4]{12} },$$
where the final simplification comes from invoking the following Lemma.\\

\begin{LEMM}~ {\rm For} ~ $0 \le x \le 1$,
$$ ~ \sqrt{2-\sqrt{4-x^2}} \le \frac{ x}{\sqrt[4]{12}}.$$
\end{LEMM}
{\bf Proof of Lemma 1:} We wish to develop an upper bound on the function $$f(x) = 2-\sqrt{4-x^2}, \quad {\rm for }~~ 0 \le x \le 1.$$ Thus our goal is to identify a function $g(x)$, where $f'(x) \le g'(x)$ for $0 \le x \le 1$, and $g(0) = f(0)$. The derivative of $f(x)$ can be upper bounded easily as follows $$f'(x) = \frac{x}{\sqrt{4-x^2}} \le \frac{x}{\sqrt{3}}, \quad {\rm for }~~ 0 \le x \le 1.$$ Thus, $g'(x) = x/\sqrt{3},$ and $g(x) = x^2/\sqrt{12}$; this ensures that $f'(x) \le g'(x)$ for $0 \le x \le 1$, and $g(0) = f(0)$. By the Fundamental Theorem of Integral Calculus, $g(x)$ provides an upper bound for $f(x)$ over the domain of interest where, $0 \le x \le 1.$ To obtain the final result, take the square root of both sides, $\sqrt{2-\sqrt{4-x^2}} \le \sqrt{x^2/\sqrt{12}} = x/{ \sqrt[4]{12}}.$ \hfill $\square$\\


Second, we will develop the lower bound on the LHS of the greedy selection criterion. To ensure that we select a point from $\set_k$ at the first iteration, we require that $y_i$'s nearest neighbor belongs to the same subspace cluster. Let $y_{nn}^i$ denote the nearest neighbor to $y_i$
$$y_{nn}^i = \arg  ~\max_{j \ne i} | \langle y_i, y_j \rangle |.$$
If $y_{nn}^i$ and $y_i$ both lie in $\set_k$, then the first point selected via OMP will result in EFS. 

Let us assume that the points in $\set_k$ admit an $\epsilon$-covering of the subspace cluster $\sub_k$, or that ${\rm cover}( \set_k) = \epsilon/2$. In this case, we have the following bound in effect
$$\max_{y_j \in \set_k}  |\langle s^{m} , y_j \rangle | \ge \sqrt{1 - \frac{\epsilon^2}{4}}.$$


Putting our upper and lower bound together and rearranging terms, we arrive at our final condition on the mutual coherence
\begin{equation*}
\mu_c < \sqrt{1 - \frac{ \epsilon^2}{4}} - \cos(\theta_0) \frac{ \epsilon}{ \sqrt[4]{12} }.
\end{equation*}
Since we have shown that this condition is sufficient to guarantee EFS at each step of Alg.\ \ref{alg:omp} provided the residual stays in the correct subspace, Thm.\ \ref{theo:EFS} follows by induction.\hfill $\square$


\subsection{Proof of Theorem \ref{theo:boundedvecs}}
\label{sec:proofboundedsubs}
%

To prove Thm.\ \ref{theo:boundedvecs}, we will assume that the union of subspaces is uniformly bounded in accordance with (\ref{eq:boundunion}). This assumption enables us to develop a tighter upper bound on the mutual coherence between any residual signal $s \in \sub_i$ and the points in $\set_j$. Since $s \in \sub_i$, the residual can be expressed as $s = \dct_i \alpha$, where $\dct_i \in \R^{n \times k_i}$ is an ONB that spans $\sub_i$ and $\alpha = \dct_i^T s$. Similarly, we can write each point in $\set_j$ as $y = \dct_j \beta$, where $\dct_j \in \R^{n \times k_j}$ is an ONB that spans $\sub_j$, $\beta = \dct_j^T y$. Let $\mathcal{B}_j = \{ \dct_j^T y_i \}_{i=1}^{d_j}$ denote the set of all subspace coefficients for all $y_i \in \set_j$. 


The coherence between the residual and a point in a different subspace can be expanded as follows:
\begin{align}
\nonumber\max_{y \in \set_j}\frac{ | \langle s, y \rangle | } { \| s\|_2 } &= \max_{\beta \in \mathcal{B}_j} \frac{ | \langle \dct_i \alpha , \dct_j \beta \rangle |}  { \| \alpha \|_2 }\\
\nonumber &= \max_{\beta \in \mathcal{B}_j} \frac{ | \langle \alpha , \dct^T_i\dct_j \beta \rangle |}   { \| \alpha \|_2 }\\
\nonumber &= \max_{\beta \in \mathcal{B}_j}\frac{|\langle \alpha , U \Sigma V^T  \beta \rangle | } { \| \alpha \|_2 }\\
\nonumber &= \max_{\beta \in \mathcal{B}_j} \frac{| \langle U^T \alpha , \Sigma V^T  \beta \rangle | } { \| \alpha \|_2 } \\
&\le \max_{\beta \in \mathcal{B}_j} \frac{\| U^T \alpha \|_{\infty}} { \|  \alpha \|_2 }   \| \Sigma V^T  \beta \|_{1},
\label{eqn:tackle}
\end{align}
where the last step comes from an application of Holder's inequality, i.e., $| \langle w, z \rangle | < \| w \|_{\infty} \| z \|_1$.

Now, we tackle the final term in (\ref{eqn:tackle}), which we can write as 
\begin{align}
\max_{\beta \in \mathcal{B}_j} \| \Sigma V^T  \beta \|_{1} &= \max_{y \in \set_j}  \|  \Sigma V^T \Phi_j^T y \|_1 = \max_{y \in \set_j}  \|  \Sigma (\Phi_j V)^T y \|_1,
\end{align}
where the matrix $\Phi_j V$ contains the principal vectors in subspace $\sub_j$. Thus, this term is simply a sum of weighted inner products between the principal vectors $\Phi_j V$ and all of the points in $\sub_j$, where $\Sigma$ contains the cross-spectra in its diagonal entries.

Since we have assumed that the union is bounded, this implies that the inner product between the first $q$ principal vectors and the points in $\set_j$ are bounded by $\gamma$, where $q = \| \sigma_{ij} \|_0 = {\rm rank}(G)$. Let $\Phi_j V_q \in \R^{n \times q}$ be the first $q$ singular vectors of $G$ corresponding to the nonzero singular values in $\Sigma$ and let $\Sigma_q \in \R^{q \times q}$ be a diagonal matrix with the first $q$ nonzero singular values of $G$ along its diagonal. It follows that $\| \Sigma (\Phi_j V)^T y \|_{\infty} = \| \Sigma_q  (\Phi_j V_q)^T y \|_{\infty} \le \gamma$. Now, suppose that the bounding constant $\gamma < \sqrt{1/q}$. In this case, 
\begin{align}
\label{eq:tracenormbound}
 \max_{y \in \set_j}  \|  \Sigma (\Phi_j V)^T y \|_1 \le \gamma \| \sigma_{ij} \|_1.
\end{align}
Note that for bounded unions of subspaces, the term on the right can be made small by requiring that the bounding constant $\gamma \ll 1$. Plugging this bound into (\ref{eqn:tackle}), we obtain the following expression
\begin{align*}
\max_{y \in \set_j} ~ \frac{ | \langle s , y \rangle | } { \| r \|_2 } & \le \gamma\| \sigma_{ij} \|_1\frac{\| U^T \alpha \|_{\infty}} { \|  \alpha \|_2 } = \gamma\| \sigma_{ij} \|_1 \| U \|_{2,2} =  \gamma\| \sigma_{ij} \|_1,
\end{align*}
where this last simplification comes from the fact that $U$ is unitary and has spectral norm equal to one. Note that this bound on the mutual coherence is informative only when $\gamma \| \sigma_{ij} \|_1 <  \sigma_{\max} \le 1$. This completes the proof. \hfill$\square$\\


\section*{Acknowledgements}
Thanks to Dr.\ Christoph Studer, Dr.\ Chinmay Hegde, and Mahdi Soltanolkotabi for helpful discussions and comments on this paper. Thanks also to Dr.\ Arian Maleki and Dr.\ Joel Tropp for helpful discussions. We would like to thank the anonymous reviewers, whose comments and suggestions were invaluable. ED was supported by a NSF GRFP 0940902 and a Texas Instruments Distinguished Graduate Fellowship. ACS and RGB were partially supported by following grants: NSF CCF-1117939, CCF-0431150, CCF-0728867, CCF-0926127; DARPA N66001-11-1-4090, N66001-11-C-4092; ONR N00014-08-1-1112, N00014-10-1-0989; AFOSR FA9550-09-1-0432; ARO MURIs W911NF-07-1-0185 and W911NF-09-1-0383.

\bibliography{csbib}

\begin{thebibliography}{34}
\providecommand{\natexlab}[1]{#1}
\providecommand{\url}[1]{\texttt{#1}}
\expandafter\ifx\csname urlstyle\endcsname\relax
  \providecommand{\doi}[1]{doi: #1}\else
  \providecommand{\doi}{doi: \begingroup \urlstyle{rm}\Url}\fi

\bibitem[Aharon et~al.(2006)Aharon, Elad, and Bruckstein]{ksvd}
M.~Aharon, M.~Elad, and A.~Bruckstein.
\newblock K-{SVD}: An algorithm for designing overcomplete dictionaries for
  sparse representation.
\newblock \emph{IEEE Trans. Signal Processing}, 54\penalty0 (11):\penalty0
  4311--4322, 2006.

\bibitem[Arias-Castro et~al.(2011)Arias-Castro, Chen, and Lerman]{locallinear}
E.~Arias-Castro, G.~Chen, and G.~Lerman.
\newblock Spectral clustering based on local linear approximations.
\newblock \emph{Electron. J. Stat.}, 5\penalty0 (0):\penalty0 217--240, 2011.

\bibitem[Baraniuk et~al.(2010)Baraniuk, Cevher, Duarte, and Hegde]{modelcs}
R.~G. Baraniuk, V.~Cevher, M.~Duarte, and C.~Hegde.
\newblock Model-based compressive sensing.
\newblock \emph{IEEE Trans. Inform. Theory}, 56\penalty0 (4):\penalty0
  1982--2001, 2010.

\bibitem[Basri and Jacobs(2003)]{basrijacobs}
R.~Basri and D.~Jacobs.
\newblock Lambertian reflectance and linear subspaces.
\newblock \emph{IEEE Trans. Pattern Anal. Machine Intell.}, 25\penalty0
  (2):\penalty0 218--233, February 2003.

\bibitem[Blumensath and Davies(2009)]{Blu}
T.~Blumensath and M.~Davies.
\newblock Sampling theorems for signals from the union of finite-dimensional
  linear subspaces.
\newblock \emph{IEEE Trans. Inform. Theory}, 55\penalty0 (4):\penalty0
  1872--1882, 2009.

\bibitem[Chen and Lerman(2009)]{chen2009scc}
G.~Chen and G.~Lerman.
\newblock Spectral curvature clustering.
\newblock \emph{Int. J. Computer Vision}, 81:\penalty0 317--330, 2009.

\bibitem[Chen et~al.(1998)Chen, Donoho, and Saunders]{DonohoBP}
S.~Chen, D.~Donoho, and M.~Saunders.
\newblock Atomic decomposition by basis pursuit.
\newblock \emph{SIAM J. Sci. Comp.}, 20\penalty0 (1):\penalty0 33--61, 1998.

\bibitem[Davis et~al.(1994)Davis, Mallat, and Zhang]{DavisOMP}
G.~Davis, S.~Mallat, and Z.~Zhang.
\newblock Adaptive time-frequency decompositions.
\newblock \emph{SPIE J. Opt. Engin.}, 33\penalty0 (7):\penalty0 2183--2191,
  1994.

\bibitem[Dyer(2011)]{DyerMS}
E.~L. Dyer.
\newblock Endogenous sparse recovery.
\newblock Master's thesis, Electrical \& Computer Eng. Dept., Rice University,
  2011.

\bibitem[Dyer et~al.(2010)Dyer, Duarte, Johnson, and Baraniuk]{dyerlva}
E.~L. Dyer, M.~Duarte, D.~J. Johnson, and R.~G. Baraniuk.
\newblock Recovering spikes from noisy neuronal calcium signals via structured
  sparse approximation.
\newblock \emph{Proc. Int. Conf. on Latent Variable Analysis and Sig.
  Separation}, pages 604--611, 2010.

\bibitem[Elhamifar and Vidal(2009)]{elhamifar2009sparse}
E.~Elhamifar and R.~Vidal.
\newblock Sparse subspace clustering.
\newblock In \emph{Proc. IEEE Conf. Comp. Vis. Patt. Recog. (CVPR)}, June 2009.

\bibitem[Elhamifar and Vidal(2010)]{vidaldisjoint}
E.~Elhamifar and R.~Vidal.
\newblock Clustering disjoint subspaces via sparse representation.
\newblock In \emph{Proc. IEEE Int. Conf. Acoust., Speech, and Signal Processing
  (ICASSP)}, pages 1926--1929, March 2010.

\bibitem[Elhamifar and Vidal(2013)]{vidaljournal}
E.~Elhamifar and R.~Vidal.
\newblock Sparse subspace clustering: algorithm, theory, and applications.
\newblock \emph{IEEE Trans. Pattern Anal. Machine Intell.}, 2013.

\bibitem[Georghiades et~al.(2001)Georghiades, Belhumeur, and Kriegman]{yaleb}
A.~S. Georghiades, P.~N. Belhumeur, and D.~J. Kriegman.
\newblock From few to many: Illumination cone models for face recognition under
  variable lighting and pose.
\newblock \emph{IEEE Trans. Pattern Anal. Machine Intell.}, 23\penalty0
  (6):\penalty0 643--660, 2001.

\bibitem[Gowreesunker et~al.(2011)Gowreesunker, Tewfik, Tadipatri, Ashe,
  Pellize, and Gupta]{gowree2011neuraldecode}
B.~V. Gowreesunker, A.~Tewfik, V.~Tadipatri, J.~Ashe, G.~Pellize, and R.~Gupta.
\newblock A subspace approach to learning recurrent features from brain
  activity.
\newblock \emph{IEEE Trans. Neur. Sys. Reh}, 19\penalty0 (3):\penalty0
  240--248, 2011.

\bibitem[Kanatani(2001)]{kanatani01}
K.~Kanatani.
\newblock Motion segmentation by subspace separation and model selection.
\newblock In \emph{Proc. IEEE Int. Conf. Comp. Vis. (ICCV)}, 2001.

\bibitem[Lu and Do(2008)]{unionsubs}
Y.~Lu and M.~Do.
\newblock Sampling signals from a union of subspaces.
\newblock \emph{IEEE Sig. Proc. Mag.}, 25\penalty0 (2):\penalty0 41--47, March
  2008.

\bibitem[Mailh\`{e} et~al.(2008)Mailh\`{e}, Lesage, Gribonval, Bimbot, and
  Vandergheynst]{mailhe2008}
B.~Mailh\`{e}, S.~Lesage, R.~Gribonval, F.~Bimbot, and P.~Vandergheynst.
\newblock Shift-invariant dictionary learning for sparse representations:
  extending {K}-{SVD}.
\newblock In \emph{Proc. Europ. Sig. Processing Conf. (EUSIPCO)}, 2008.

\bibitem[Mailh\`{e} et~al.(2012)Mailh\`{e}, Barchiesi, and
  Plumbley]{Mailhe2012}
B.~Mailh\`{e}, D.~Barchiesi, and M.~D. Plumbley.
\newblock {INK}-{SVD}: Learning incoherent dictionaries for sparse
  representations.
\newblock In \emph{Proc. IEEE Int. Conf. Acoust., Speech, and Signal Processing
  (ICASSP)}, pages 3573 --3576, march 2012.

\bibitem[Mairal et~al.(2008)Mairal, Bach, Ponce, Sapiro, and
  Zisserman]{mairal2008discriminative}
J.~Mairal, F.~Bach, J.~Ponce, G.~Sapiro, and A.~Zisserman.
\newblock Discriminative learned dictionaries for local image analysis.
\newblock In \emph{Proc. IEEE Conf. Comp. Vis. Patt. Recog. (CVPR)}, June 2008.

\bibitem[Ng et~al.(2002)Ng, Jordan, and Weiss]{ng2002spectral}
A.Y. Ng, M.I. Jordan, and Y.~Weiss.
\newblock On spectral clustering: Analysis and an algorithm.
\newblock \emph{Proc. Adv. in Neural Processing Systems (NIPS)}, 2:\penalty0
  849--856, 2002.

\bibitem[Olshausen and Field(1997)]{bruno97}
B.~Olshausen and D.~Field.
\newblock Sparse coding with an overcomplete basis set: a strategy employed by
  {V}1.
\newblock \emph{Vision Res.}, 37:\penalty0 3311--3325, 1997.

\bibitem[Ramamoorthi(2002)]{facesubs}
R.~Ramamoorthi.
\newblock Analytic {PCA} construction for theoretical analysis of lighting
  variability in images of a lambertian object.
\newblock \emph{IEEE Trans. Pattern Anal. Machine Intell.}, 24\penalty0
  (10):\penalty0 1322--1333, 2002.

\bibitem[Ramirez et~al.(2010)Ramirez, Sprechmann, and Sapiro]{ramirez2010}
I.~Ramirez, P.~Sprechmann, and G.~Sapiro.
\newblock Classification and clustering via dictionary learning with structured
  incoherence and shared features.
\newblock In \emph{Proc. IEEE Conf. Comp. Vis. Patt. Recog. (CVPR)}, pages
  3501--3508, June 2010.

\bibitem[Shi and Malik(2000)]{graphcuts}
J.~Shi and J.~Malik.
\newblock Normalized cuts and image segmentation.
\newblock \emph{IEEE Trans. Pattern Anal. Machine Intell.}, 22\penalty0
  (8):\penalty0 888--905, August 2000.

\bibitem[Soltanolkotabi and Cand\`{e}s(2012)]{Candes12}
M.~Soltanolkotabi and E.~J. Cand\`{e}s.
\newblock A geometric analysis of subspace clustering with outliers.
\newblock \emph{Annals of Statistics}, 40\penalty0 (4):\penalty0 2195--2238,
  2012.

\bibitem[Soltanolkotabi et~al.(2013)Soltanolkotabi, Elhamifar, and
  Cand{\`e}s]{Candes13}
M.~Soltanolkotabi, E.~Elhamifar, and E.~J. Cand{\`e}s.
\newblock Robust subspace clustering.
\newblock \emph{CoRR}, abs/1301.2603, 2013.

\bibitem[Tropp(2004)]{Tropp04}
J.~A. Tropp.
\newblock Greed is good: {A}lgorithmic results for sparse approximation.
\newblock \emph{IEEE Trans. Inform. Theory}, 50\penalty0 (10):\penalty0
  2231--2242, 2004.

\bibitem[Tropp(2006)]{Tropp04relax}
J.~A. Tropp.
\newblock Just relax: convex programming methods for identifying sparse signals
  in noise.
\newblock \emph{IEEE Trans. Inform. Theory}, 52\penalty0 (3):\penalty0 1030
  --1051, March 2006.

\bibitem[Vidal(2011)]{vidal2011spmag}
R.~Vidal.
\newblock Subspace clustering.
\newblock \emph{IEEE Sig. Proc. Mag.}, 28\penalty0 (2):\penalty0 52--68, 2011.

\bibitem[Vidal et~al.(2005)Vidal, Ma, and Sastry]{GPCA2005}
R.~Vidal, Y.~Ma, and S.~Sastry.
\newblock Generalized principal component analysis ({GPCA}).
\newblock \emph{IEEE Trans. Pattern Anal. Machine Intell.}, 27\penalty0
  (12):\penalty0 1945--1959, 2005.

\bibitem[Wang and Xu(2013)]{Xu2013}
Y.~Wang and H.~Xu.
\newblock Noisy sparse subspace clustering.
\newblock \emph{Proc. Int. Conf. Machine Learning}, 2013.

\bibitem[Yan and Pollefeys(2006)]{LSA}
J.~Yan and M.~Pollefeys.
\newblock A general framework for motion segmentation: Independent,
  articulated, rigid, non-rigid, degenerate and non-degenerate.
\newblock In \emph{Proc. European Conf. Comp. Vision (ECCV)}, 2006.

\bibitem[Zhang et~al.(2012)Zhang, Szlam, Wang, and Lerman]{zhang2010hybrid}
T.~Zhang, A.~Szlam, Y.~Wang, and G.~Lerman.
\newblock Hybrid linear modeling via local best-fit flats.
\newblock \emph{Int. J. Computer Vision}, 100\penalty0 (3):\penalty0 217--240,
  2012.

\end{thebibliography}

\end{document}